\documentclass[letterpaper]{article} 
\usepackage{aaai23}  
\usepackage{times}  
\usepackage{helvet}  
\usepackage{courier}  
\usepackage[hyphens]{url}  
\usepackage{graphicx} 
\usepackage{natbib}  
\usepackage{caption} 
\usepackage{microtype}
\usepackage{booktabs} 

\usepackage{subfigure}
\usepackage{relsize}
\usepackage{multirow}
\usepackage{threeparttable}
\usepackage{stmaryrd}
\usepackage{longtable}
\usepackage{wrapfig,lipsum}

\usepackage{algorithm}
\usepackage{algorithmic}

\usepackage{amsmath,amsfonts,bm}

\def\eqref#1{equation~\ref{#1}}

\def\1{\bm{1}}

\def\re{{\mathsf{e}}}

\def\rv{{\mathsf{v}}}

\def\rx{{\mathsf{x}}}
\def\ry{{\mathsf{y}}}

\def\rvz{{\mathbf{z}}}

\def\vtheta{{\bm{\theta}}}

\def\vw{{\bm{w}}}

\DeclareMathAlphabet{\mathsfit}{\encodingdefault}{\sfdefault}{m}{sl}
\SetMathAlphabet{\mathsfit}{bold}{\encodingdefault}{\sfdefault}{bx}{n}

\def\sS{{\mathbb{S}}}
\def\sT{{\mathbb{T}}}

\def\sV{{\mathbb{V}}}

\def\sX{{\mathbb{X}}}
\def\sY{{\mathbb{Y}}}

\renewcommand{\b}[1]{\mathbf{#1}}

\newcommand{\stitle}[1]{\vspace{1ex}\noindent{\bf #1}}

\usepackage{newfloat}
\usepackage{listings}
\DeclareCaptionStyle{ruled}{labelfont=normalfont,labelsep=colon,strut=off} 
\floatstyle{ruled}
\newfloat{listing}{tb}{lst}{}
\floatname{listing}{Listing}
\title{Time Series Contrastive Learning with Information-Aware Augmentations}
\author{
    Dongsheng Luo\textsuperscript{\rm 1},
    Wei Cheng\textsuperscript{\rm 2},
    Yingheng Wang\textsuperscript{\rm 4},
    Dongkuan Xu\textsuperscript{\rm 3},
    Jingchao Ni\textsuperscript{\rm 2},
    Wenchao Yu\textsuperscript{\rm 2},
    Xuchao Zhang\textsuperscript{\rm 2},
    Yanchi Liu\textsuperscript{\rm 2},
    Yuncong Chen\textsuperscript{\rm 2},
    Haifeng Chen\textsuperscript{\rm 2},
    Xiang Zhang\textsuperscript{\rm 4}
}
\affiliations{
    \textsuperscript{\rm 1}Florida International University, 
    \textsuperscript{\rm 2}NEC Labs America, 
    \textsuperscript{\rm 3}North Carolina State University,
    \textsuperscript{\rm 4}The Pennsylvania State University\\

    dluo@flu.edu,\{weicheng,wyu,yanchi,yuncong,haifeng\}@nec-labs.com, yw2349@cornell.edu, dxu27@ncsu.edu,\\jingchni@amazon.com, xuchaozhang@microsoft.com,xzz89@psu.edu
}

\usepackage{bibentry}

\begin{document}

\maketitle

\begin{abstract}
Various contrastive learning approaches have been proposed in recent years and achieve significant empirical success. While effective and prevalent, contrastive learning has been less explored for time series data. A key component of contrastive learning is to select appropriate augmentations imposing some priors to construct feasible positive samples, such that an encoder can be trained to learn robust and discriminative representations. Unlike image and language domains where ``desired'' augmented samples can be generated with the rule of thumb guided by prefabricated human priors, the ad-hoc manual selection of time series augmentations is hindered by their diverse and human-unrecognizable temporal structures. How to find the desired augmentations of time series data that are meaningful for given contrastive learning tasks and datasets remains an open question. In this work, we address the problem by encouraging both high \textit{fidelity} and \textit{variety} based upon information theory. A theoretical analysis leads to the criteria for selecting feasible data augmentations. On top of that, we propose a new contrastive learning approach with information-aware augmentations, InfoTS, that adaptively selects optimal augmentations for time series representation learning. Experiments on various datasets show highly competitive performance with up to 12.0\%  reduction in MSE on forecasting tasks and up to 3.7\% relative improvement in accuracy on classification tasks over the leading baselines.
\end{abstract}

\section{Introduction}

Time series data in the real world is high dimensional, unstructured, and complex with unique properties, leading to challenges for data modeling~\citep{yang200610}. In addition, without human recognizable patterns, it is much harder to label time series data than images and languages in real-world applications. These labeling limitations hinder deep learning methods, which typically require a huge amount of labeled data for training, being applied on time series data~\citep{eldele2021time}. Representation learning learns a fixed-dimension embedding from the original time series that keeps its inherent features. Compared to the raw time series data,  these representations are with better transferability and generalization capacity. To deal with labeling limitations, contrastive learning methods have been widely adopted in various domains for their soaring performance on representation learning, including vision, language, and graph-structured data~\citep{chen2020simple,xie2019unsupervised,you2020graph}. In a nutshell, contrastive learning methods typically train an encoder to map instances to an embedding space where dissimilar (negative) instances are easily distinguishable from similar (positive) ones and model predictions to be invariant to small noise applied to either input examples or hidden states. 

\begin{figure*}
    \centering
    \includegraphics[width=5in]{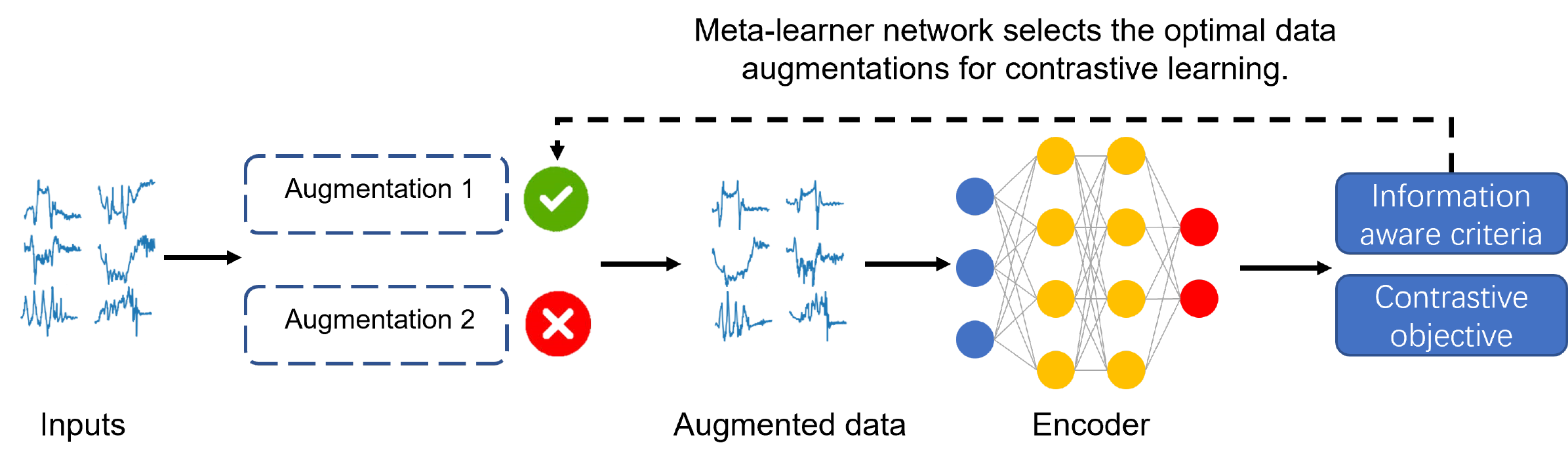}
    \caption{InfoTS is composed of three parts: (1) candidate transformation that generates different augmentations of the original inputs, (2) a meta-learner network that selects the optimal augmentations, (3) an encoder that learns representations of time series instances. The meta-learner is learned in tandem with contrastive encoder learning. }\vspace{-0.6cm}
    \label{fig:framework}
\end{figure*}

Despite being effective and prevalent, contrastive learning has been less explored in the time series domain~\citep{eldele2021time,franceschi2019unsupervised,fan2020self,tonekaboni2021unsupervised}. Existing contrastive learning approaches often involve a specific data augmentation strategy that creates novel and realistic-looking training data without changing its label to construct positive alternatives for any input sample. Their success relies on carefully designed rules of thumb guided by domain expertise. Routinely used data augmentations for contrastive learning are mainly designed for image and language data, such as color distortion, flip, word replacement, and back-translation~\citep{chen2020simple,luo2021unsupervised}. These augmentation techniques generally do not apply to time series data. Recently, some researchers propose augmentations for time series to enhance the size and quality of the training data~\citep{TSAugmentationSurvey}. For example, TS-TCC~\citep{eldele2021time} and Self-Time~\citep{fan2020self} adopt jittering, scaling, and permutation strategies to generate augmented instances. Franceschi et.al. propose to extract subsequences for data augmentation~\citep{franceschi2019unsupervised}. In spite of the current progress, existing methods have two main limitations. First, unlike images with human recognizable features, time series data are often associated with inexplicable underlying patterns. Strong 
augmentation such as permutation may ruin such patterns and consequently, the model will mistake the negative handcrafts for positive ones. While weak augmentation methods such as jittering may generate augmented instances that are too similar to the raw inputs to be informative enough for contrastive learning. On the other hand, time series datasets from different domains may have diverse natures. Adapting a universal data augmentation method, such as subsequence~\citep{xie2019unsupervised}, for all datasets and tasks leads to sub-optimal performances. Other works follow empirical rules to select suitable augmentations from expensive trial-and-error. Akin to hand-crafting features, hand-picking choices of data augmentations are undesirable from the learning perspective. The diversity and heterogeneity of real-life time series data further hinder these methods away from wide applicability.

To address the challenges, we first introduce the criteria for selecting good data augmentations in contrastive learning. Data augmentation benefits generalizable, transferable, and robust representation learning by correctly extrapolating the input training space to a larger region~\citep{wilk2018learning}. The positive instances enclose a discriminative zone in which all the data points should be similar to the original instance. 
The desired data augmentations for contrastive representation learning should have both \textit{high fidelity} and \textit{high variety}. High fidelity encourages the augmented data to maintain the semantic identity that is invariant to transformations~\citep{wilk2018learning}. For example, if the downstream task is classification, then the generated augmentations of inputs should be class-preserving.
Meanwhile, generating augmented samples with high variety benefits representation learning by increasing the generalization capacity~\citep{chen2020simple}. From the motivation, we theoretically analyze the information flows in data augmentations based on information theory and derive the criteria for selecting desired time series augmentations. Due to the inexplicability in practical time series data, we assume that the semantic identity is presented by the target in the downstream task. Thus, high fidelity can be achieved by maximizing the mutual information between the downstream label and the augmented data. A one-hot pseudo label is assigned to each instance in the unsupervised setting when downstream labels are unavailable. These pseudo-labels encourage augmentations of different instances to be distinguishable from each other. We show that data augmentations preserving these pseudo labels can add new information without decreasing the fidelity. 
Concurrently, we maximize the entropy of augmented data conditional on the original instances to increase the variety of data augmentations.

Based on the derived criteria, we propose an adaptive data augmentation method, InfoTS (as shown in Figure~\ref{fig:framework}), to avoid ad-hoc choices or painstaking trial-and-error tuning. Specifically, we utilize another neural network, denoted by meta-learner, to learn the augmentation prior in tandem with contrastive learning. The meta-learner 
automatically selects optimal augmentations from candidate augmentations to generate feasible positive samples. Along with randomly sampled negative instances, augmented instances are then fed into a time series encoder to learn representations in a contrastive manner. With a reparameterization trick, the meta-learner can be efficiently optimized with back-propagation based on the proposed criteria. Therefore, the meta-learner can automatically select data augmentations in a per dataset and per learning task manner without resorting to expert knowledge or tedious downstream validation. Our main contributions include:

\begin{itemize}
    \item We propose criteria to guide the selection of data augmentations for contrastive time series representation learning without prefabricated knowledge.
    \item We propose an approach to automatically select feasible data augmentations for different time series datasets, which can be efficiently optimized with back-propagation.
    \item We empirically verify the effectiveness of the proposed criteria to find optimal data augmentations. Extensive experiments demonstrate that InfoTS can achieve highly competitive performance with up to 12.0\% reduction in MSE on forecasting tasks and up to
    3.7\% relative improvement in accuracy on classification tasks over the leading baselines. 
\end{itemize}
\section{Methodology}
\label{sec:method}
\subsection{Notations and Problem Definition}
A time series instance $x$ has dimension $T \times F$, where $T$ is the length of sequence and $F$ is the dimension of features.  Given a set of time series instances $\sX$, we aim to learn an encoder $f_{\vtheta}(x)$ that maps each instance $x$ to a fixed-length vector $\b{z}\in\mathbb{R}^D$, where $\vtheta$ is the learnable parameters of the encoder network and  $D$ is the dimension of representation vectors. In semi-supervised settings, each instance $x$ in the labelled set $\sX_{L} \subseteq  \sX$ is associated with a label $y$ for the downstream task. Specially, $\sX_{L} = \sX$ holds in the fully supervised setting. In the work, we use the Sans-serif style lowercase letters, such as $\rx$, to denote random time series variables and italic lowercase letters, such as $x$, for sampled instances.

\subsection{Information-Aware Criteria for Good Augmentations}
\label{sec:criteria}
\begin{figure}[!t]
    \centering
    \subfigure[Information-aware criteria]{\includegraphics[width=1.5in]{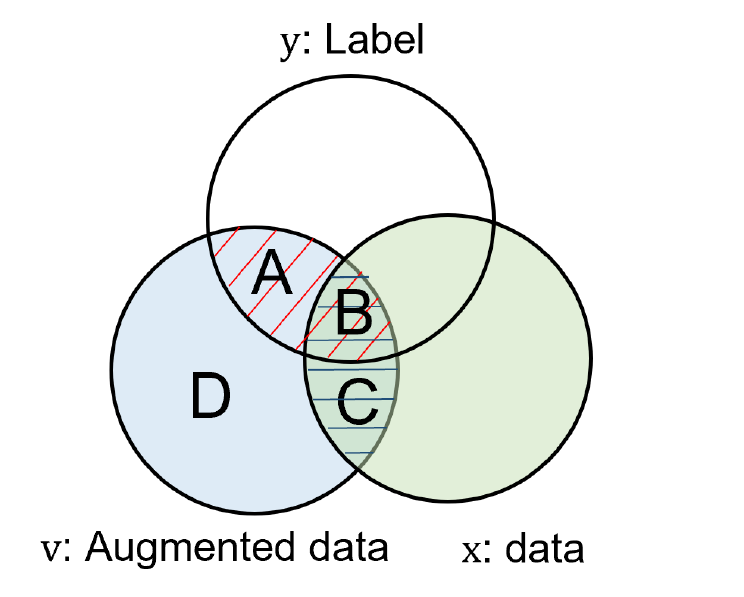}\label{fig:indc:xyv}} \hspace{1.5cm}
    \subfigure[Examples]{\includegraphics[width=2.2in]{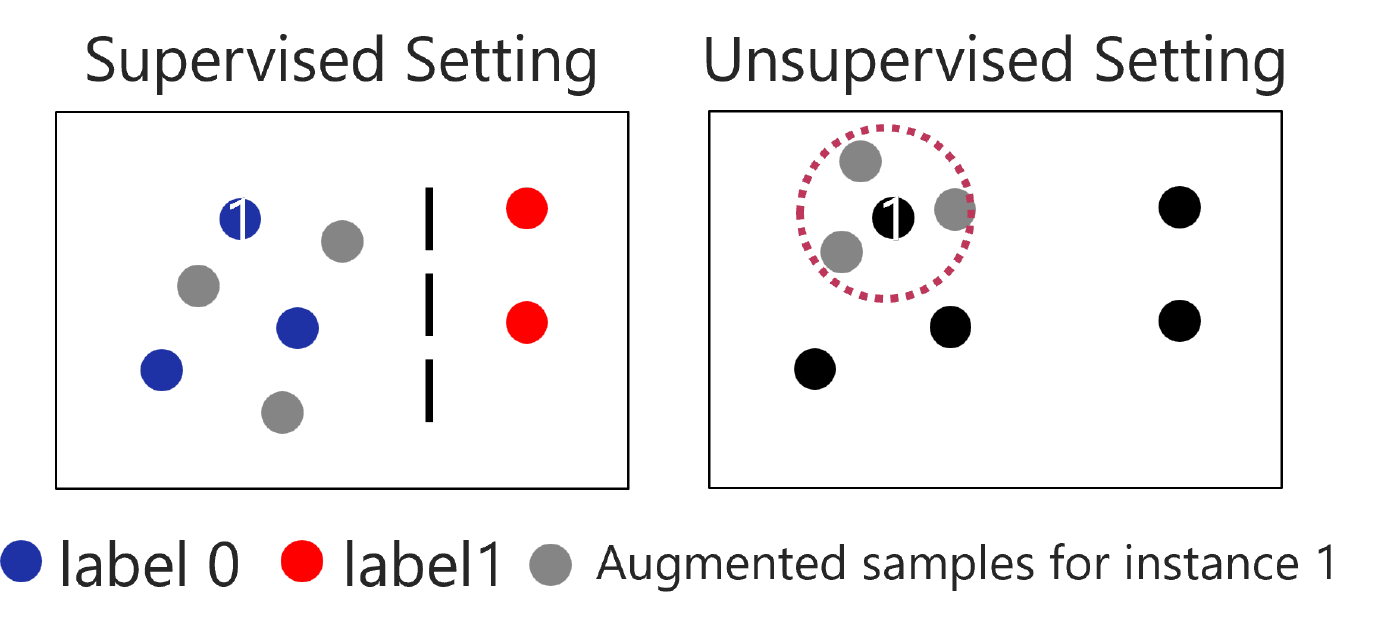}\label{fig:indc:augmentation}}
        \vspace{-0.32cm}
    \caption{Illustration of the criteria. (a) The proposed criteria have two components: high fidelity, and variety. Fidelity is represented by the area of A+B, the mutual information between augmented data $\rv$ and label $\ry$. Variety is denoted by A+D, the entropy of $\rv$ conditioned on the raw input $\rx$. (b) In the supervised setting, good data augmentations generate instances in the area constrained by the label to enlarge the input training space. In the unsupervised setting, with one-hot-based pseudo labels, the generated instances are constrained to the region around the raw input. Such that they are still distinguishable from other instances. }
    \label{fig:criteria}
        \vspace{-0.32cm}
\end{figure}

The goal of data augmentation for contrastive learning is to create realistically rational instances that maintain semantics through different transformation approaches. Unlike instances in vision and language domains, the underlying semantics of time series data is not recognizable to human, making it hard, if not impossible, to include human knowledge to data augmentation for time series data.  For example, rotating an image will not change its content or the label. While permuting a time series instance may ruin its signal patterns and generates a meaningless time series instance. In addition, the tremendous heterogeneity of real-life time series datasets further makes selections based on trial-and-errors impractical.  Although multiple data augmentation methods have been proposed for time series data, there is less discussion on what is a good augmentation that
is meaningful for a given learning task and dataset without prefabricated human priors. From our perspective, ideal data augmentations for contrastive representation should keep high fidelity, high variety, and adaptive to different datasets.  The illustration and examples are shown in Figure~\ref{fig:criteria}.

\stitle{High Fidelity.} Augmentations with high fidelity maintain the semantic identity that is invariant to transformations. Considering the inexplicability in practical time series data, it is challenging to visually check the fidelity of augmentations. Thus, we assume that the semantic identity of a time series instance is presented by its label in the downstream task, which might be either available or unavailable during the training period. Here, we start our analysis from the supervised case and will extend it to the unsupervised case later. Inspired by on the information bottleneck~\citep{tishby2000information}, we define the objective that keeps high fidelity as the large mutual information (MI) between augmentation $\rv$ and the label $\ry$, i.e., $\text{MI}(\rv;\ry)$.

We consider augmentation $\rv$ as a \textit{probabilistic} function of $\rx$ and a random variable $\b{\epsilon}$, that $\rv=g(\rx;\epsilon)$. From the definition of mutual information, we have $\text{MI}(\rv;\ry)=H(\ry)-H(\ry|\rv)$, where $H(\ry)$ is the (Shannon) entropy of $\ry$ and $H(\ry|\rv)$ is the entropy of $\ry$ conditioned on augmentation $\rv$. Since $H(\ry)$ is irrelevant to data augmentations, the objective is equivalent to minimizing the conditional entropy $H(\ry|\rv)$.
Considering the efficient optimization, we follow~\citep{ying2019gnnexplainer} and ~\citep{luo2020parameterized} to approximate it with  cross-entropy  between $\ry$ and $\hat{\ry}$, where $\hat{\ry}$ is the prediction with augmentation $\rv$ as the input and calculated via
\begin{equation}
       \rv = g(\rx;\epsilon) \quad \quad  \rvz = f_{\vtheta}(\rv) \quad \quad \hat{\ry}=h_{\vw}(\rvz),
\end{equation}
 where $\mathbf{z}$ is the representation and $h_{\vw}(\cdot)$ is a prediction projector parameterized by $\vw$. The prediction projector is optimized by the classification objective. Then, the objective of high fidelity for supervised or semi-supervised cases is to minimize 
 \begin{equation}
 \vspace{-0.1cm}
 \label{eq:supervisedfidelity}
      \text{CE} (\ry;\hat{\ry}) =  -\sum_{c=1}^C P(\ry=c)\log P(\hat{\ry}=c),
 \end{equation}
 \vspace{-0.1cm}
where $C$ is the number of labels.

In the \textit{unsupervised} settings where $\ry$ is unavailable, \textit{one-hot} encoding $\ry_s \in \mathbb{R}^{|\mathbb{X}|} $ is utilized as the pseudo label to replace $\ry$ in Eq.~(\ref{eq:supervisedfidelity}).
The motivation is that augmented instances are still distinguishable from other instances with the classifier. We theoretically show that augmentations that preserving pseudo labels have the following properties.

\noindent{\textbf{Property 1}\emph { (Preserving Fidelity).}} \emph{If  augmentation $\rv$ preserves the one-hot encoding pseudo label, the mutual information between $\rv$ and the downstream task label $\ry$ (although not visible to training) is equivalent to that between raw input $\rx$ and $\ry$, i.e., $\text{MI}(\rv;\ry) = \text{MI} (\rx;\ry)$.}

\noindent{\textbf{Property 2}\emph { (Adding New Information).}} \emph{By preserving the one-hot encoding pseudo label, augmentation $\rv$ contains new information comparing to the raw input $\rx$, i.e., $H(\rv)\geq H(\rx)$.}

Detailed proofs are shown in the Appendix~\ref{sec:app:proof}. These properties show that in the unsupervised setting, preserving the one-hot encoding pseudo label guarantees that the generated augmentations will not decrease the fidelity, regardless of the downstream tasks and variances inherent in the augmentations. Concurrently, it may introduce new information for contrastive learning.

Since the number of labels is equal to the number of instances in dataset $\mathbb{X}$ in an unsupervised case, direct optimization of Eq.~(\ref{eq:supervisedfidelity}) is inefficient and unscalable. Thus, we further relax it by approximating $\ry$ with the batch-wise one-hot encoding $\ry_B$, which decreases the number of labels $C$ from the dataset size to the batch size.

\stitle{High Variety.} Sufficient variances in augmentations improve the generalization capacity of contrastive learning models.  In the information theory, the uncertainty inherent in the random variable's possible outcomes is described by its entropy. Considering that augmented instances are generated based on the raw input $\rx$, we maximize the entropy of $\rv$ conditioned on $\rx$, $H(\rv|\rx)$, to maintain a high variety of augmentations. From the definition of conditional entropy, we have
\begin{equation}
    H(\rv|\rx) = H(\rv)-\text{MI}(\rv;\rx).
\end{equation}
We dismiss the first part since the unconstrained entropy of $\rv$ can be dominated by meaningless noise.  Considering the continuity of both $\rv$ and $\rx$, we minimize the mutual information between $\rv$ and $\rx$ by minimize the leave-one-out upper (L1Out) bound~\citep{poole2019variational}. 
Other MI upper bounds, such as contrastive log-ratio upper bound of mutual information~\citep{cheng2020club}, can also conveniently be the plug-and-play component in our framework. Then, the objective to encourage high variety is to minimize the L1Out between $\rv$ and $\rx$:
\begin{equation}
\label{eq:variety}
\text{I}_\text{L1Out}(\rv;\rx) = \mathbb{E}_x \left[ \log \frac{\exp(\text{sim}(\mathbf{z}_{x},\mathbf{z}_v))}{\sum_{x'\in \mathbb{X},x'\neq x} \exp(\text{sim}(\mathbf{z}_x,\mathbf{z}_{v'}))} \right],
\end{equation}
where $v'$ is an augmented instance of input instance $x'$.  $\b{z}_x$, $\b{z}_v$, and $\b{z}_{v'}$ are representations of instance $x$, $v$, and $v'$ respectively. $\text{sim}(\b{z}_1,\b{z}_2) = \b{z}_1^T\b{z}_2$ is the inner product of vectors $\b{z}_1$ and $\b{z}_2$.

\stitle{Criteria.} Combining the information-aware definition of both high fidelity and variety, we propose the criteria for selecting good augmentations without prior knowledge,
\begin{equation}
\label{eq:criteria}
    \min_{\rv} \text{I}_\text{L1Out}(\rv;\rx)+\beta \text{CE}(\ry;h_\vw(f_\vtheta(\rv))),
\end{equation}
where $\beta$ is a hyper-parameter to achieve the trade-off between fidelity and variety. Note that in the \textit{unsupervised} settings, $\ry$ is replaced by \textit{one-hot} encoding pseudo label.

\stitle{Relation to Information Bottleneck.}
Although the formation is similar to information bottleneck in data compression, $ \min _{p(\re|\rx)} \text{MI}(\rx;\re)-\beta \text{MI}(\re;\ry)$, our criteria are different in the following aspects. 
First, $\re$ in the information bottleneck is a representation of input $\rx$, while $\rv$ in Eq.(\ref{eq:criteria}) represents the augmented instances. Second, the information bottleneck aims to keep minimal and sufficient information for data compression, while our criteria are designed for data augmentations in contrastive learning.  Third, in the information bottleneck, the compressed representation $\re$ is a deterministic function of input $\rx$ with no variances. $\text{MI}(\re;\ry)$ and $\text{MI}(\re;\rx)$ are constraint by $\text{MI}(\rx;\ry)$ and $H(\rx)$ that $\text{MI}(\re;\ry) \leq \text{MI}(\rx;\ry)$ and  $\text{MI}(\re;\rx) = H(\re)$, where $H(\re)$ is the entropy of $\re$. In our criteria, 
$\rv$ is a probabilistic function of input $\rx$. As a result, the variances of $\rv$ make the augmentation space much larger than the compression representation space in the information bottleneck.

\stitle{Relation to InfoMin.}
InfoMin is designed based on the information bottleneck that good views should keep minimal and sufficient information from the original input~\citep{tian2020makes}. Similar to the information bottleneck, InfoMin assumes that augmented views are functions of the input, which heavily constrains the variance of data augmentations. Besides, high fidelity property is dismissed in the unsupervised setting. It works for image datasets due to the availability of human knowledge. However, it may fail to generate reasonable augmentations for time series data. In addition, they adopt adversarial learning, which minimizes a lower bound of MI, to increase the variety of augmentations. While to minimize statistical dependency, we prefer an upper bound, such as L1Out, instead of lower bounds.

\subsection{Time Series Meta-Contrastive Learning}
We aim to design a learnable augmentation selector that learns to select feasible augmentations in a data-driven manner. With such adaptive data augmentations, the contrastive loss is then used to train the encoder that learns representations from raw time series. 

\subsubsection{Architecture}
\label{sec:method:arch}
The adopted encoder $f_\vtheta(x): \mathbb{R}^{T\times F}\rightarrow \mathbb{R}^{D}$ consists of two components, a fully connected layer, and a 10-layer dilated CNN module~\citep{franceschi2019unsupervised,yue2021learning}.  To explore the inherent structure of time series, we include both global-wise (instance-level) and local-wise (subsequence-level) losses in the contrastive learning framework to train the encoder.

\stitle{Global-wise contrastive loss} is designed to capture the instance level relations in a time series dataset. Formally, given a batch of time series instances $\mathbb{X}_B \subseteq \mathbb{X}$, for each instance $x \in \mathbb{X}_B$, we generate an augmented instance $v$ with an adaptively selected transformation, which will be introduced later. $(x,v)$ is regarded as a positive pair and other $(B-1)$ combinations $\{(x,v')\}$, where $v'$ is an augmented instance of $x'$ and $x'\neq x$, are considered as negative pairs. 
Following~\citep{chen2020simple,you2020graph}, we design the global-wise contrastive loss based on InfoNCE~\citep{hjelm2018learning}. The batch-wise instance-level contrastive loss is
\begin{equation}
\label{eq:global}
    \mathcal{L}_g = -\frac{1}{|\sX_B|}\sum_{x\in \mathbb{X}_B} \log \frac{\exp(\text{sim}(\mathbf{z}_x,\mathbf{z}_v))}{\sum_{x'\in \mathbb{X}_B} \exp(\text{sim}(\mathbf{z}_x,\mathbf{z}_{v'}))}.
\end{equation}

\stitle{Local-wise contrastive loss } is proposed to explore the intra-temporal relations in time series. For an augmented instance $v$ of a time series instance $x$, we first split it into a set of subsequences $\sS$, each with length $L$.  For each subsequence $s\in \sS$, we follow \citep{tonekaboni2021unsupervised} to generate a positive pair $(s,p)$ by selecting another subsequence close to it. Non-neighboring samples, $\bar{\mathcal{N}}_s$, are adopted to generate negative pairs. Detailed descriptions can be found in Appendix~\ref{sec:app:algorithm}.
Then, the local-wise contrastive loss for an instance $x$ is:

\begin{small}
\begin{equation}
\label{eq:local}
    \mathcal{L}c_x = -\frac{1}{|\sS|}\sum_{s \in \sS} \log \frac{\exp(\text{sim}(\mathbf{z}_s,\mathbf{z}_p))}{\exp(\text{sim}(\mathbf{z}_s,\mathbf{z}_p))+\sum_{j\in \bar{\mathcal{N}}_s} \exp(\text{sim}(\mathbf{z}_s,\mathbf{z}_j))}.
\end{equation}
\end{small}
\normalfont
Across all instances in a batch, we have $\mathcal{L}_c = \frac{1}{|\sX_B|}\sum_{x\in \mathbb{X}_B} \mathcal{L}c_x$. The final contrastive objective is:
\begin{equation}
\label{eq:contrastive}
    \min_\vtheta \mathcal{L}_g +\alpha \mathcal{L}_c,
\end{equation}
where $\alpha$ is a hyper-parameter to achieve the trade-off between global and local contrastive losses.

\subsubsection{Meta-learner Network}
Previous time series contrastive learning methods~\citep{franceschi2019unsupervised,fan2020self,eldele2021time,tonekaboni2021unsupervised} generate augmentations with either rule of thumb guided by prefabricated human priors or tedious trial-and-errors, which are designed for specific datasets and learning tasks.
In this part, we discuss how to adaptively select the optimal augmentations with a meta-learner network based on the proposed information-aware criteria. We can regard its choice of optimal augmentation as a kind of prior selection. 
We first choose a set of candidate transformations $\sT$, such as jittering and time warping. Each candidate transformation $t_i\in \sT$ is associated with a weight $p_i \in (0,1)$, inferring the probability of selecting transformation $t_i$. For an instance $x$, the augmented instance $v_i$ through transformation $t_i$ can be computed by:
\begin{equation}
\label{eq:bernoulli}
    a_i \sim \text{Bernoulli}(p_i) \quad \quad v_i = (1-a_i)x+a_it_i(x).
\end{equation}
Considering multiple transformations, we pad all $v_i$ to be with the same length. Then, the adaptive augmented instance can be achieved by combining candidate ones, $v=\frac{1}{|\sT|}\sum_i v_i$.

To enable the efficient optimization with gradient-based methods, we approximate discrete Bernoulli processes with binary concrete distributions~\citep{maddison2016concrete}.  Specifically, we approximate $a_i$ in Eq.~(\ref{eq:bernoulli}) with 
\begin{equation}
\centering
\begin{aligned}
\label{eq:concrete}
    &\epsilon \sim \text{Uniform}(0,1) \\
    & a_i=\sigma((\log \epsilon - \log (1-\epsilon) + \log\frac{p_i}{1-p_i})/\tau),
\end{aligned}
\end{equation}
where $\sigma(\cdot)$ is the sigmoid function and $\tau$ is the temperature  controlling the approximation.  The rationality of such approximation is given in Appendix~\ref{sec:app:proof}. Moreover, with temperature $\tau >0 $, the gradient $\frac{\partial v }{\partial p_i}$ is well-defined. Therefore, our meta-network is end-to-end differentiable. Detailed algorithm is shown in Appendix~\ref{sec:app:algorithm}.

\section{Related Work}
\label{relatedwork}
\subsection{Contrastive Time Series Representation Learning}
Contrastive learning has been utilized widely in representation learning with superior performances in various domains~\citep{chen2020simple,xie2019unsupervised,you2020graph}. Recently, some efforts have been devoted to applying contrastive learning to the time series domain~\citep{oord2018representation,franceschi2019unsupervised,fan2020self,eldele2021time,tonekaboni2021unsupervised,yue2021learning}. Time Contrastive Learning trains a feature extractor with a multinomial logistic regression classifier to discriminate all segments in a time series~\citep{hyvarinen2016unsupervised}.  In \citep{franceschi2019unsupervised}, Franceschi et.al. generate positive and negative pairs based on subsequences.  TNC employs a debiased contrastive objective to  ensure that in the representation space, signals in the local neighborhood are distinguishable from non-neighboring signals~\citep{tonekaboni2021unsupervised}. SelfTime adopts multiple hand-crafted augmentations for unsupervised time series contrastive learning by exploring both inter-sample and intra-sample relations~\citep{fan2020self}. TS2Vec learns a representation for each time stamp and conducts contrastive learning in a hierarchical way~\citep{yue2021learning}. However, data augmentations in these methods are either universal or selected by error-and-trail, hindering them away from been widely applied in complex real-life datasets.

\subsection{Time Series Forecasting} Forecasting is a critical task in time series analysis. Deep learning architectures used in the literature include Recurrent Neural Networks (RNNs)~\citep{salinas2020deepar,oreshkin2019n}, Convolutional Neural Networks (CNNs)~\citep{bai2018empirical}, Transformers~\citep{li2019enhancing,zhou2021informer}, and Graph Neural Networks (GNNs)~\citep{cao2021spectral}. N-BEATS deeply stacks fully-connected layers with backward and forward residual links for univariate times series forecasting~\citep{oreshkin2019n}. TCN utilizes a deep CNN architecture with dilated causal convolutions~\citep{bai2018empirical}. Considering both long-term dependencies and short-term trends in multivariate time series, LSTnet combines both CNNs and RNNS in a unified model~\citep{lai2018modeling}. LogTrans brings the Transformer model to time series forecasting with causal convolution in its attention mechanism~\cite{li2019enhancing}. Informer further proposes a sparse self-attention mechanism to reduce the time complexity and memory usage~\citep{zhou2021informer}. StemGNN is a GNN based model that considers the intra-temporal and inter-series correlations simultaneously~\cite{cao2021spectral}. Unlike these works, we aim to learn general representations for time series data that can not only be used for forecasting but also other tasks, such as classification. Besides, the proposed framework is compatible with various architectures as encoders.

\subsection{Adaptive Data Augmentation}
Data augmentation is an important component in contrastive learning. Existing researches reveal that the choices of optimal augmentation are dependent on downstream tasks and datasets~\citep{chen2020simple,fan2020self}. Some researchers have explored adaptive selections of optimal augmentations for contrastive learning in the vision field. 
AutoAugment automatically searches the combination of translation policies via a reinforcement learning method~\citep{cubuk2019autoaugment}. Faster-AA improves the searching pipeline for data augmentation using a differentiable policy network~\citep{hataya2020faster}. DADA further introduces an unbiased gradient estimator for an efficient one-pass optimization strategy~\citep{li2020dada}. Within contrastive learning frameworks, Tian et.al. apply the Information Bottleneck theory that optimal views should share minimal and sufficient information, to guide the selection of good views for contrastive learning in the vision domain~\citep{tian2020makes}.
Considering the inexplicability of time series data, directly applying the InfoMin framework may keep insufficient information during augmentation. Different from \citep{tian2020makes}, we focus on the time series domain and propose an end-to-end differentiable method to automatically select the optimal augmentations for each dataset.

\section{Experiments}
\label{sec:exp}
In this section, we compare InfoTS with SOTA baselines on time series forecasting and classification tasks. We also conduct case studies to show insights into the proposed criteria and meta-learner network. Detailed experimental setups are shown in Appendix~\ref{sec:app:setup}. Full experimental results and extra experiments, such as parameter sensitivity studies, are presented in  Appendix~\ref{sec:app:exp}.

\begin{table*}[h!]
  \centering
    \caption{Univariate time series forecasting results.}
  \label{tab:forecast-univar}
  \scalebox{0.8}{
  \begin{tabular}{lccccccccccccccc}
  \toprule
         &  & \multicolumn{2}{c}{InfoTS} &
                  \multicolumn{2}{c}{TS2Vec} &
         \multicolumn{2}{c}{Informer} &
         \multicolumn{2}{c}{LogTrans} &
         \multicolumn{2}{c}{N-BEATS} &
         \multicolumn{2}{c}{TCN} &
         \multicolumn{2}{c}{LSTnet} \\
    \cmidrule(r){3-4} \cmidrule(r){5-6} \cmidrule(r){7-8} \cmidrule(r){9-10} \cmidrule(r){11-12} \cmidrule(r){13-14}  \cmidrule(r){15-16}
    Dataset & $L_y$ & MSE & MAE & MSE & MAE & MSE & MAE & MSE & MAE & MSE & MAE & MSE & MAE & MSE & MAE \\
    \midrule
    \multirow{5}*{ETTh$_1$}
    & 24 & \textbf{0.039} & \textbf{0.149} & \textbf{0.039} & 0.152 & 0.098 & 0.247 & 0.103 & 0.259 & 0.094 & 0.238 & 0.075 & 0.210 & 0.108 & 0.284 \\
    & 48 &\textbf{0.056} & \textbf{0.179} & 0.062 & 0.191 & 0.158 & 0.319 & 0.167 & 0.328 & 0.210 & 0.367 & 0.227 & 0.402 & 0.175 & 0.424 \\
    & 168 &\textbf{0.100} & \textbf{0.239} & 0.134 & 0.282 & 0.183 & 0.346 & 0.207 & 0.375 & 0.232 & 0.391 & 0.316 & 0.493 & 0.396 & 0.504 \\
    & 336 &\textbf{0.117} & \textbf{0.264} & 0.154 & 0.310 & 0.222 & 0.387 & 0.230 & 0.398 & 0.232 & 0.388 & 0.306 & 0.495 & 0.468 & 0.593 \\
    & 720 & \textbf{0.141} & \textbf{0.302} & 0.163 & 0.327 & 0.269 & 0.435 & 0.273 & 0.463 & 0.322 & 0.490 & 0.390 & 0.557 & 0.659 & 0.766 \\
    \midrule

    \multirow{5}*{ETTh$_2$}
    & 24  &\textbf{0.081} &\textbf{0.215} & 0.090 & 0.229 & 0.093 & 0.240 & 0.102 & 0.255 & 0.198 & 0.345 & 0.103 & 0.249 & 3.554 & 0.445 \\
    & 48 & \textbf{0.115} & \textbf{0.261} & 0.124 & 0.273 & 0.155 & 0.314 & 0.169 & 0.348 & 0.234 & 0.386 & 0.142 & 0.290 & 3.190 & 0.474 \\
    & 168  &\textbf{0.171}& \textbf{0.327} & 0.208 & 0.360 & 0.232 & 0.389 & 0.246 & 0.422 & 0.331 & 0.453 & 0.227 & 0.376 & 2.800 & 0.595 \\
    & 336 & \textbf{0.183}& \textbf{0.341} & 0.213 & 0.369 & 0.263 & 0.417 & 0.267 & 0.437 & 0.431 & 0.508 & 0.296 & 0.430 & 2.753 & 0.738 \\
    & 720  &\textbf{0.194}& \textbf{0.357} & 0.214 & 0.374 & 0.277 & 0.431 & 0.303 & 0.493 & 0.437 & 0.517 & 0.325 & 0.463 & 2.878 & 1.044 \\
    
    \midrule

    \multirow{5}*{ETTm$_1$}
    & 24 & \textbf{0.014} &\textbf{0.087} & 0.015 &  0.092 & 0.030 & 0.137 & 0.065 & 0.202 & 0.054 & 0.184 & 0.041 & 0.157 & 0.090 & 0.206   \\
    & 48 & \textbf{0.025} & \textbf{0.117}  &  0.027 &  0.126 & 0.069 & 0.203 & 0.078 & 0.220 & 0.190 & 0.361 & 0.101 & 0.257 & 0.179 & 0.306 \\
    & 96 & \textbf{0.036} &\textbf{0.142}  &  0.044 &  0.161 & 0.194 & 0.372 & 0.199 & 0.386 & 0.183 & 0.353 & 0.142 & 0.311 & 0.272 & 0.399 \\
    & 288 &\textbf{0.071}& \textbf{0.200} &  0.103 &  0.246 & 0.401 & 0.554 & 0.411 & 0.572 & 0.186 & 0.362 & 0.318 & 0.472 & 0.462 & 0.558 \\
    & 672 & \textbf{0.102}& \textbf{0.240}  & 0.156 &  0.307 & 0.512 & 0.644 & 0.598 & 0.702 & 0.197 & 0.368 & 0.397 & 0.547 & 0.639 & 0.697 \\

    \midrule
  
    \multirow{5}*{Electricity}
    & 24 & \textbf{0.245} & \textbf{0.269} & 0.260 & 0.288 & 0.251 & 0.275 & 0.528 & 0.447 & 0.427 & 0.330 & 0.263 & 0.279 & 0.281 & 0.287 \\
    & 48 & \textbf{0.294} & \textbf{0.301} & 0.319 & 0.324 & 0.346 & 0.339 & 0.409 & 0.414 & 0.551 & 0.392 & 0.373 & 0.344 & 0.381 & 0.366 \\
    & 168 & \textbf{0.402} & \textbf{0.367} & 0.427 & 0.394 & 0.544 & 0.424 & 0.959 & 0.612 & 0.893 & 0.538 & 0.609 & 0.462 & 0.599 & 0.500 \\
    & 336 & \textbf{0.533} & \textbf{0.453} & 0.565 & 0.474 & 0.713 & 0.512 & 1.079 & 0.639 & 1.035 & 0.669 & 0.855 & 0.606 & 0.823 & 0.624 \\

    \midrule
 
    \multicolumn{2}{l}{Avg.} & \textbf{0.154} & \textbf{0.253}  & 0.175 &0.278 & 0.263 & 0.367 & 0.336 & 0.419 & 0.338 & 0.402 & 0.289 & 0.359 & 1.090 & 0.516 \\
    \bottomrule
  \end{tabular}
  }
\end{table*}

\begin{table*}[h]
  \centering
    \caption{Multivarite time series classification on 30 UEA datasets.}\label{tab:full-uea-unsup-sum}
    \scalebox{0.8}{
  \begin{tabular}{lccccccccccc}
  \toprule
    Method  & $\text{InfoTS}_s$ & InfoTS & TS2Vec & T-Loss & TNC & TS-TCC & TST & DTW \\
    \midrule
    Avg. ACC & \textbf{0.730}& 0.714& 0.704& 0.658 & 0.670 & 0.668 & 0.617 & 0.629 \\
    Avg. Rank & \textbf{1.967}& 2.633& 3.067& 3.833 & 4.367 & 4.167 & 5.0 & 4.366 \\
    \bottomrule
  \end{tabular}
 } \vspace{-0.5cm}
\end{table*}

\subsection{Time Series Forecasting}
 Time series forecasting aims to predict the future $L_y$ time stamps, with the last $L_x$ observations. We follow~\citep{yue2021learning} to train a linear model regularized with the L2 norm penalty to make predictions. The output has dimension $L_y$ in the univariate case and $L_y\times F$ for the multivariate case, where $F$ is the feature dimension.

 \stitle{Datasets and Baselines.} Four benchmark datasets for time series forecasting are adopted, including ETTh1, ETTh2, ETTm1~\citep{zhou2021informer}, and the Electricity dataset~\citep{Dua2019}. These datasets are used in both univariate and multivariate settings.  We compare unsupervised InfoTS to the SOTA baselines, including TS2Vec~\citep{yue2021learning}, Informer~\citep{zhou2021informer}, StemGNN~\citep{cao2021spectral},  TCN~\citep{bai2018empirical}, LogTrans~\citep{li2019enhancing}, LSTnet~\citep{lai2018modeling}, and N-BEATS~\citep{oreshkin2019n}. Among these methods, N-BEATS is merely designed for the univariate and StemGNN is for multivariate only. We refer to~\citep{yue2021learning} to set up baselines for a fair comparison. Standard metrics for a regression problem, Mean Squared Error (MSE), and Mean Absolute Error (MAE) are utilized for evaluation. Evaluation results of univariate time series forecasting are shown in Table \ref{tab:forecast-univar}, while multivariate forecasting results are reported in the Appendix~\ref{sec:app:full} due to the space limitation. 

\stitle{Performance.} As shown in Tabel~\ref{tab:forecast-univar} and Tabel~\ref{tab:forecast-multivar}, comparison in both univariate and multivariate settings indicates that InfoTS consistently matches or outperforms the leading baselines. Some results of StemGNN are unavailable due to the out-of-memory issue~\citep{yue2021learning}. Specifically, we have the following observations. TS2Vec, another contrastive learning method with data augmentations, achieves the second-best performance in most cases. The consistent improvement of TS2Vec over other baselines indicates the effectiveness of contrastive learning for time series representations learning.  However, such universal data augmentations may not be the most informative ones to generate positive pairs. Comparing to TS2Vec, InfoTS decreases the average MSE by 12.0\%, and the average MAE by 9.0\% in the univariate setting. In the multivariate setting, the MSE and MAE decrease by 5.5\% and 2.3\%, respectively. The reason is that InfoTS can adaptively select the most suitable augmentations in a data-driven manner with high variety and high fidelity. Encoders trained with such informative augmentations learn representations with higher quality. 

\normalsize
\begin{figure}[h!]
    \centering
    \subfigure[MSE]{\includegraphics[width=1.3in]{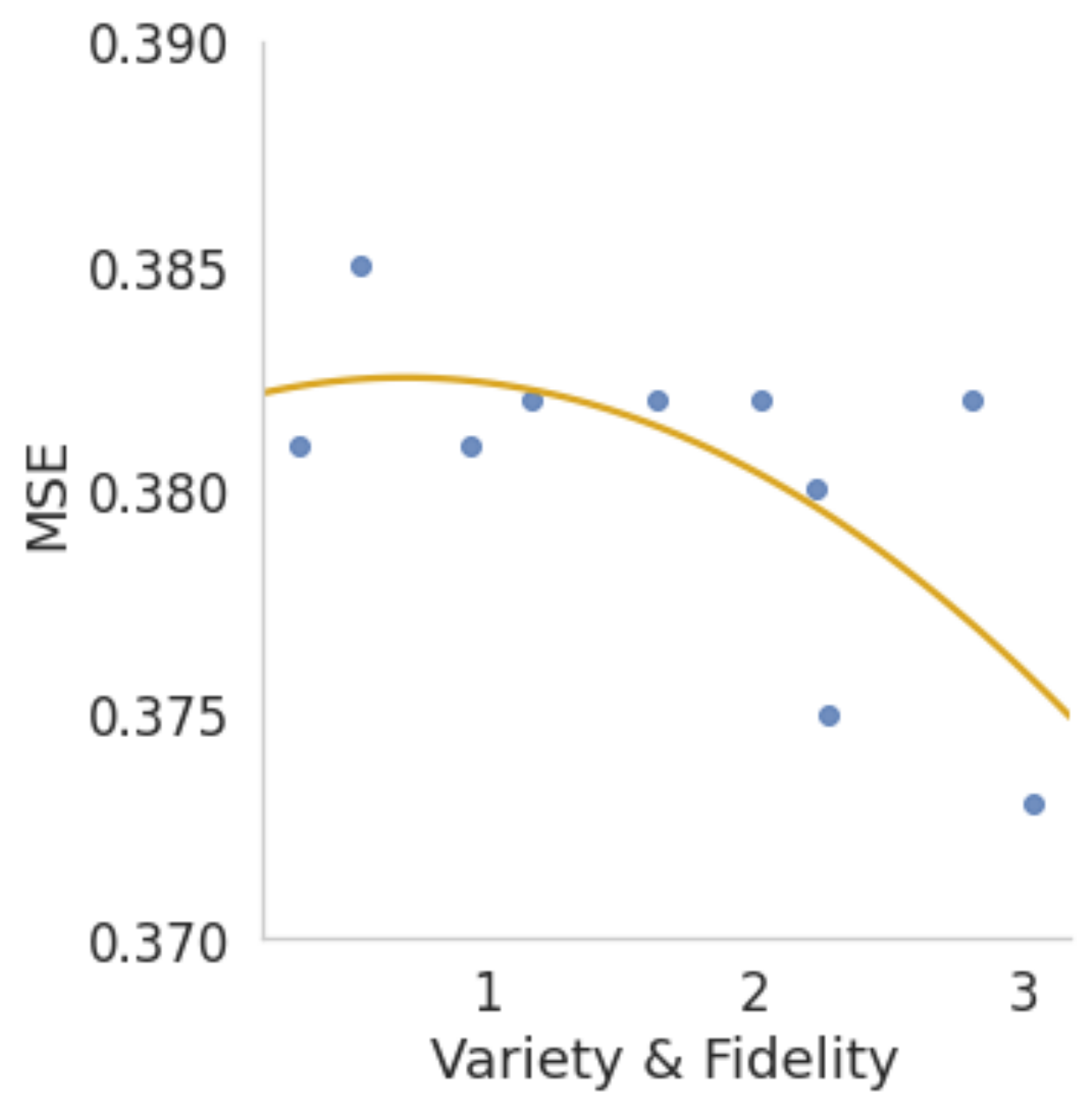}\label{fig:app:criteria:mse}}\quad\quad\quad\quad
    \subfigure[MAE]{\includegraphics[width=1.3in]{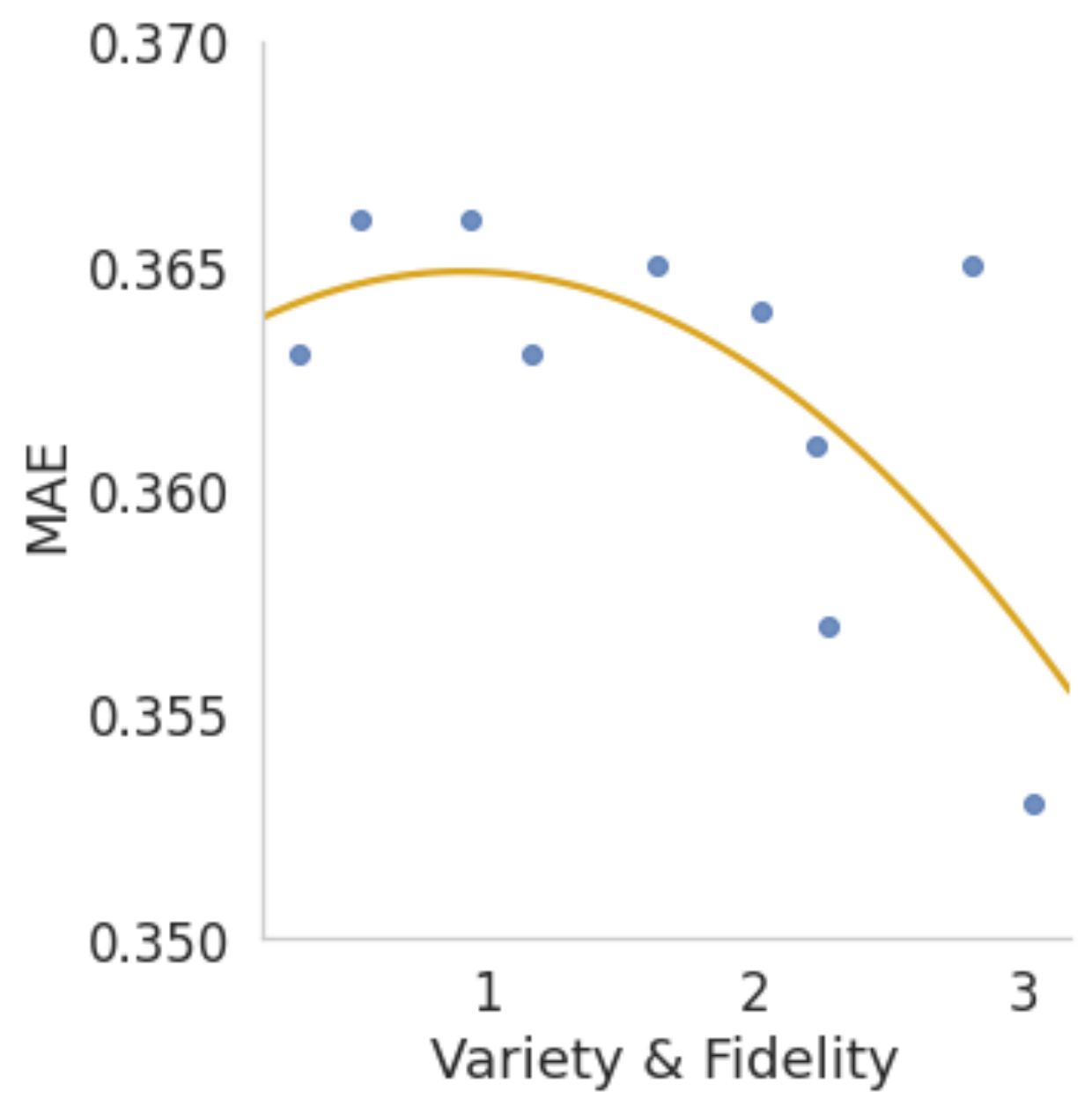}\label{fig:app:criteria:mae}}\vspace{-0.2cm}
    \caption{Evaluation of the criteria on forecasting.}
    \label{fig:exp:criteria}
    \vspace{-0.5cm}
\end{figure}

\subsection{Time Series Classification}
\label{sec:exp:classifcation}
Following the previous setting, we evaluate the quality of representations on time series classification in a standard supervised manner~\citep{franceschi2019unsupervised,yue2021learning}. We train an SVM classifier with a radial basis function kernel on top of representations in the training split and then compare the prediction in the test set.

\stitle{Datasets and Baselines.} Two types of benchmark datasets are used for evaluations. The UCR archive~\citep{dau2019ucr} is a collection of 128 univariate time series datasets and the UEA archive~\citep{bredin2017tristounet} consists of 30 multivariate datasets. We compare InfoTS with baselines including TS2Vec~\citep{yue2021learning}, T-Loss~\citep{franceschi2019unsupervised}, TS-TCC~\citep{eldele2021time}, TST~\citep{zerveas2021transformer}, and DTW~\citep{franceschi2019unsupervised}. For our methods, $\text{InfoTS}_s$, training labels are only used to train the meta-learner network to select suitable augmentations, and InfoTS is with a purely unsupervised setting for representation learning.

\stitle{Performance.} The results on the UEA datasets are summarized in Table~\ref{tab:full-uea-unsup-sum}. Full results are provided in Appendix~\ref{sec:app:full}.  With the ground-truth label guiding the meta-learner network, InfoTS$_s$ substantially outperforms other baselines. On average, it improves the classification accuracy by 3.7\% over the best baseline, TS2Vec, with an average rank value 1.967 on all 30 UEA datasets.  Under the purely unsupervised setting, InfoTS preserves fidelity by adopting one-hot encoding as the pseudo labels. InfoTS achieves the second best average performance in Table~\ref{tab:full-uea-unsup-sum}, with an average rank value 2.633. Performances on the 128 UCR datasets are shown in Table~\ref{tab:app:full-ucr} in Appendix~\ref{sec:app:full}. These datasets are univariate with easily recognized patterns, where data augmentations have marginal or even negative effects~\citep{yue2021learning}. However, with adaptively selected augmentations for each dataset based on our criteria, InfoTS$_s$ and InfoTS still outperform the state-of-the-arts.

\subsection{Ablation Studies}
To present deep insights into the proposed method, we conduct multiple ablation studies on the Electricity dataset to empirically verify the effectiveness of the proposed information aware criteria and the framework to adaptively select suitable augmentations.  MSE is utilized for evaluation. 

\begin{table}[h]
  \centering
    \caption{Ablation studies on Electricity with MSE as the evaluation.}
  \label{tab:exp:ablation}
  \scalebox{0.8}{
  \begin{tabular}{lccccc}
  \toprule
         &\multirow{2}*{InfoTS} &
         \multicolumn{2}{c}{Data Augmentation} &
         \multicolumn{2}{c}{Meta Objective}\\
        \cmidrule(r){3-4} \cmidrule(r){5-6}
         & &
         Random &
         All  &
         w/o Fidelity&
        w/o Variety\\
     \midrule
    24  & \textbf{0.245} & 0.252 & 0.249  &0.254  &0.251 \\
    48  & \textbf{0.295} & 0.303 & 0.303  &0.306  &0.297 \\
    168 & \textbf{0.400} & 0.414 & 0.414  &0.414  &0.409 \\
    336 & \textbf{0.541} & 0.565 & 0.563  &0.562  &0.542 \\
     \midrule
    Avg. &\textbf{0.370} &0.384  &0.382	  &0.384  &0.375 \\
    \bottomrule
  \end{tabular}}
  \vspace{-0.4cm}
\end{table}

\stitle{Evaluation of The Criteria.}
In Section~\ref{sec:criteria}, we propose information-aware criteria of data augmentations for time series that good augmentations should have high variety and fidelity. With L1Out and cross-entropy as approximations, we get the criteria in Eq.~(\ref{eq:criteria}).  To empirically verify the effectiveness of the proposed criteria, we adopt two groups of augmentations, subsequence augmentations with different lengths and jitter augmentations with different standard deviations. Subsequence augmentations work on the temporal dimension, and jitter augmentations work on the feature dimension. For the subsequence augmentations, we range the ratio of subsequences $r$ in the range $[0.01,0.99]$. The subsequence augmentation with ratio $r$ is denoted by Sub$_r$, such as Sub$_{0.01}$. For the jitter augmentations, the standard deviations are chosen from the range $[0.01,3.0]$. The jitter augmentation with standard deviation $std$  is denoted by Jitter$_{std}$, such as Jitter$_{0.01}$.

Intuitively, with $r$ increasing, Sub$_r$ generates augmented instances with lower variety and higher fidelity.  For example, with $r=0.01$, Sub$_r$ generates subsequences that only keep $1\%$ time stamps from the original input, leading to  high  variety but extremely low fidelity. Similarly, for jitter augmentations, with $std$ increasing, Jitter$_{std}$ generates augmented instances with higher variety but lower fidelity.

In Figure~\ref{fig:exp:criteria}, we show the relationship between forecasting performance and our proposed criteria. In general, performance is positively related to the proposed criteria in both MAE and MSE settings, verifying the correctness of using the criteria as the objective in the meta-learner network training.

\stitle{Evaluation of The Meta-Learner Network.}
In this part, we empirically show the advantage of the developed meta-learner network on learning optimal augmentations. Results are shown in Table~\ref{tab:exp:ablation}. 
We compare InfoTS with variants ``Random'' and ``All''.  ``Random'' randomly selects an augmentation from candidate transformation functions each time and ``All'' sequentially applies transformations to generate augmented instances. Their performances are heavily affected by the low-quality candidate augmentations, verifying the key role of adaptive selection in our method. 2) To show the effects of variety and fidelity objectives in meta-learner network training, we include two variants, ``w/o Fidelity'' and ``w/o Variety'', which dismiss the fidelity or variety objective, respectively. The comparison between InfoTS and the variants empirically confirms both variety and fidelity are important for data augmentation in contrastive learning.

\section{CONCLUSIONS}
We propose an information-aware criteria of data augmentations for time series data that good augmentations should preserve high variety and high fidelity.  We approximate the criteria with a mutual information neural estimation and cross-entropy estimation. Based on the approximated criteria, we adopt a meta-learner network to adaptively select optimal augmentations for contrastive representation learning. Comprehensive experiments show that representations produced by our method are high qualified and easy to use in various downstream tasks, such as time series forecasting and classification, with state-of-the-art performances.

\bibliography{aaai23}
\clearpage
  \appendix

\section{Detailed Proofs}
\label{sec:app:proof}
\stitle{Properties of Data Augmentations that Preserve Pseudo Labels.}

We assume that $v_i$ and $v_j$ are two augmented instances of inputs $x_i$ and $x_j$, respectively.  Preserving pseudo labels defined in one-hot encoding requires that the map between variable $\rx$ and $\rv$ is one-to-many. Formally, $x_i \neq x_j \rightarrow v_i \neq v_j$. This can be proved by contradiction. If we have a pair of $(i,j)$ that $i\neq j$ and $v_i = v_j$, then we have $f_\vtheta(v_i)=f_\vtheta(v_j)$, showing that augmentations cannot preserve pseudo labels.

\noindent{\textbf{Property 1}\emph { (Preserving Fidelity).}} \emph{If  augmentation $\rv$ preserves the one-hot encoding pseudo label, the mutual information between $\rv$ and downstream task label $\ry$ (although not visible to training) is equivalent to that between raw input $\rx$ and $\ry$, i.e., $\text{MI}(\rv;\ry) = \text{MI} (\rx;\ry)$.}

\noindent{\emph {Proof.}} From the definition of mutual information, we have
\begin{equation*}
    \begin{aligned}
        \text{MI}(\rv;\ry) &= H(\ry)-H(\ry|\rv) \\
         &=  H(\ry) + \sum_{v,y} p(v,y)\log\frac{p(v,y)}{p(v)} \\
         &=  H(\ry) + \sum_{x,y} \sum_{v\in \sV(x)}p(v,y)\log\frac{p(v,y)}{p(v)},\\
    \end{aligned}
\end{equation*}
where $\sV(x)$ is the set of augmented instances of a time series instance $x$.
In the unsupervised setting where the ground-truth label $y$ is unknown, we assume that the augmentation $v$ is a (probabilistic) function of $x$ only. The only qualifier means $p(v|x,y)=p(v|x).$ Since the mapping from $\rx$ to $\rv$ is one-to-many. For each $v \in \sV(x)$ we have $p(v,y) = p(v,x,y)$ and $p(v) = p(v|x)p(x)$. 
Thus, we have 
\begin{equation*}
    \begin{aligned}
        \frac{p(v,y)}{p(v)} &= \frac{p(v,x,y)}{p(v|x)p(x)} =  \frac{p(v|x,y) p(x,y)}{p(v|x)p(x)}\\
        &= \frac{p(v|x) p(x,y)}{p(v|x)p(x)}  = \frac{p(x,y)}{p(x)}\\
         \text{MI}(\rv;\ry) &  = H(\ry) + \sum_{x,y} \sum_{v\in \sV(x)}p(v,y)\log\frac{p(x,y)}{p(x)} \\
         & =H(\ry) + \sum_{x,y}  [\sum_{v\in \sV(x)}p(v,y)]\log\frac{p(x,y)}{p(x)} \\
         & =H(\ry) + \sum_{x,y} p(x,y)\log\frac{p(x,y)}{p(x)} \\
         & = \text{MI}(\rx;\ry).
    \end{aligned}
\end{equation*}

\noindent{\textbf{Property 2}\emph { (Adding New Information).}} \emph{By preserving the one-hot encoding pseudo label, augmentation $\rv$ contains new information comparing to the raw input $\rx$, i.e., $H(\rv)\geq H(\rx)$.}

In information theory, entropy describes the amount of information of a random variable.
For simplicity, we assume a finite number of augmented instances for each input, and each augmented instance is generated independently.  Then, we have $p(x)=\sum_{ v\in \sV(x)}p(v)$. Then we have that the entropy of variable $\rx$ is no larger than the entropy of $\rv$.  
\begin{equation*}
    \begin{aligned}
        H(\rx) &= -\sum_{x} p(x)\log p(x) \\
        &= -\sum_{x} [\sum_{ v\in \sV(x)}p(v)] \log[\sum_{ v\in \sV(x)}p(v)]\\
                &= -\sum_{x} \sum_{ v\in \sV(x)}p(v) \log[\sum_{ v\in \sV(x)}p(v)]\\
        &\leq  -\sum_{x}  \sum_{ v\in \sV(x)} p(v) \log p(v)\\
        &= -\sum_{v}p(v) \log p(v)=H(\rv)
    \end{aligned}
\end{equation*}

\stitle{Rationality of Approximation of Bernoulli Distribution with Binary Concrete Distribution in Eq.~(\ref{eq:concrete}).} 

In the binary concrete distribution, parameter $\tau$ controls the temperature that achieves the trade-off between binary output and continuous optimization.  When $\tau\to0$, we have  $\lim_{\tau\to0}P(a_i=1)=p_i$, which is equivalent to the Bernoulli distribution. 

\noindent{\emph {Proof.}} 
\begin{equation*}
\begin{aligned}
        &\lim_{\tau\to0}P(a_i=1) \\
        &= \lim_{\tau\to0}P(\sigma((\log \epsilon - \log (1-\epsilon) + \log\frac{p_i}{1-p_i})/\tau)=1)\\
        & =P(\log \epsilon - \log (1-\epsilon) + \log\frac{p_i}{1-p_i} >0)\\
        &= P(\log \frac{\epsilon}{1-\epsilon} - \log\frac{1-p_i}{p_i} >0)\\
        &= P(\frac{\epsilon}{1-\epsilon}>\frac{1-p_i}{p_i})
\end{aligned}
\end{equation*}
Since $\epsilon$, and $p_i$ are both in $(0,1)$, and function $\frac{x}{1-x}$ are monotonically increasing in this region. Thus, we have 
\begin{equation*}
\begin{aligned}
        \lim_{\tau\to0}P(a_i=1) & = P(\epsilon>1-p_i)=p_i
\end{aligned}
\end{equation*}

\section{Algorithms}
\label{sec:app:algorithm}

\subsection{Training Algorithm}
\label{sec:app:algorithm:training}
The training algorithm of InfoTS under both supervised and unsupervised settings is described in Algorithm~\ref{alg:training}.  We first randomly initiate parameters in the encoder, meta-learner network, and classifier (line 2).
Given a batch of training instances $ \sX_B \subseteq  \sX$, for each candidate transformation $t_i$, we utilize binary concrete distribution to get parameters $a_i$ (lines 6-7), which indicates whether the transformation should be applied (line 8). $ \sV_B^{(i)}$ denotes the batch of augmented instances generated from transformation function $t_i$. The final augmented instances are generated by adaptively considering all candidates transformations (line 10). Parameters $\vtheta$ in the encoder are updated by minimizing the contrastive objective (lines 11-13). Meta-learner network is then optimized with information-aware criteria (lines 14-16). Then, classifier $h_\vw$ is optimized with the classification objective (line 17).

\begin{algorithm}[h]
\begin{small}
    \centering
    \caption{Algorithm for InfoTS in both supervised and unsupervised settings}
    \label{alg:training}
    \begin{algorithmic}[1]
        \STATE {\bfseries Input:}   time series dataset $ \sX$, the label set $ \sY$ (supervised setting), a set of candidate transformations $\sT$, hyper-parameters $\alpha$ and $\beta$, 
        \STATE Initialize the encoder $f_\vtheta$, parameters in meta-learner network $\{q_i\}_{i=1}^{|\sT|}$, and the classifier $h_\vw$.
        \FOR{each epoch}
            \FOR{each training batch $ \sX_B \subseteq  \sX$}
                \FOR{each transformation $t_i \in \sT$}
                    \STATE $p_i \leftarrow \sigma(q_i)$
                    \STATE $a_i \leftarrow \text{binaryConcrete}(p_i)$
                    \STATE $\sV_B^{(i)} \leftarrow $ transform each $x \in \sX_B$ with Eq.~(\ref{eq:bernoulli})
                \ENDFOR
                \STATE Get $\sV_B $ by averaging $\{\sV_B^{(i)}\}_{i=1}^{|\sT|}$.
                \STATE Compute global contrastive loss $\mathcal{L}_g$ with Eq.~(\ref{eq:global})
                \STATE Compute local contrastive loss $\mathcal{L}_c$ with Eq.~(\ref{eq:local})
                \STATE  Update parameters $\vtheta$ in the encoder with Eq.~(\ref{eq:contrastive})
                \STATE Compute fidelity loss with Eq.~(\ref{eq:supervisedfidelity})
                \STATE Compute variety loss with  Eq.~(\ref{eq:variety})
                \STATE Update parameters $\{q_i\}_{i=1}^{|\sT|}$ in the meta-learner network with Eq.~(\ref{eq:criteria})
                \STATE Update parameters $\vw$ in the classifier $h_\vw$ with the cross-entropy loss
            \ENDFOR
    \ENDFOR
    \end{algorithmic}
\end{small}
\end{algorithm}

\subsection{Implementation of Local-Wise Contrastive}
\label{sec:app:algorithm:local}
\begin{figure*}
    \centering
    \includegraphics[width=5in]{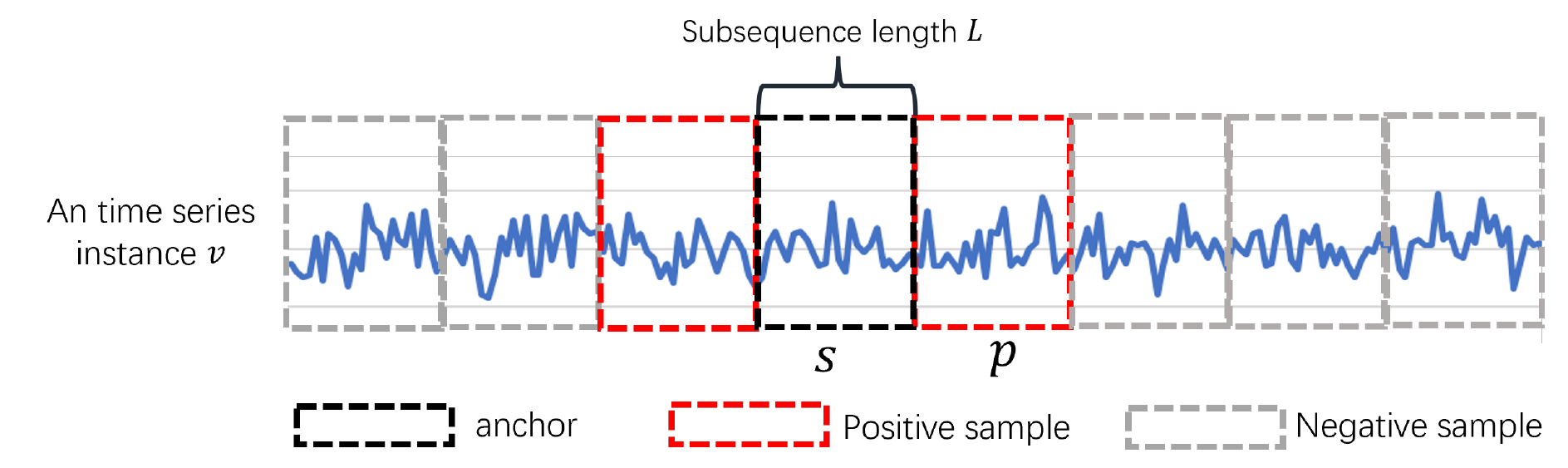}
    \caption{Positive and negative samples for a subsequence $s$.}
    \label{fig:localloss}
\end{figure*}

Local-wise contrastive loss aims to capture the intra-temporal relations in each time series instance. For an augmented instance $v$, we first split it into multiple subsequences, as shown in Figure~\ref{fig:localloss}. Each subsequence has length $L$. For each subsequence $s$, the neighboring subsequences within window size 1 are considered as positive samples. If $s$ locates at the end of $v$, then we choose the subsequence in front of $s$ as the positive pair $p$.  Otherwise, we choose the subsequence following $s$ instead.  Subsequences out of window size 1 are considered as negative samples. 

\section{Experimental Settings}
\label{sec:app:setup}
\subsection{Data Augmentations}
We follow~\citep{fan2020self} to set up candidate data augmentations, including jittering, scaling, cutout, time warping, window slicing, window warping and subsequence augmentation~\citep{franceschi2019unsupervised}. Detailed descriptions are listed as follows. 
\begin{itemize}
    \item \textit{Jittering} augmentation adds the random noise sampled from a Gaussian distribution $\mathcal{N}(0, 0.3)$ to the input time series.
    \item \textit{Scaling} augmentation multiplies the input time series by a scaling factor sampled from a Gaussian distribution $\mathcal{N}(0, 0.5)$.
    \item  \textit{Cutout} operation replaces features of 10\% randomly sampled time stamps of the input with zeros.
    \item \textit{Time warping} random changes the speed of the timeline\footnote{\url{https://tsaug.readthedocs.io/}}. The number of speed changes is 100 and the maximal ratio of max/min speed is 10. If necessary, over-sampling or sampling methods are adopted to ensure the length of the augmented instance is the same as the original one.
    \item \textit{Window slicing} randomly crops half the input time series and then linearly interpolates it back to the original length~\citep{le2016data}.
    \item \textit{Window warping} first randomly selects 30\% of the input time series along the timeline and then warps the time dimension by 0.5 or 2.  Finally, we adopt linear interpolation to transform it back to the original length \citep{le2016data}.
    \item \textit{Subsequence} operation random selects a subsequence from the input time series~\citep{yue2021learning}.
\end{itemize}

\begin{figure}
    \centering
    \subfigure[Original]{\includegraphics[width=1.3in]{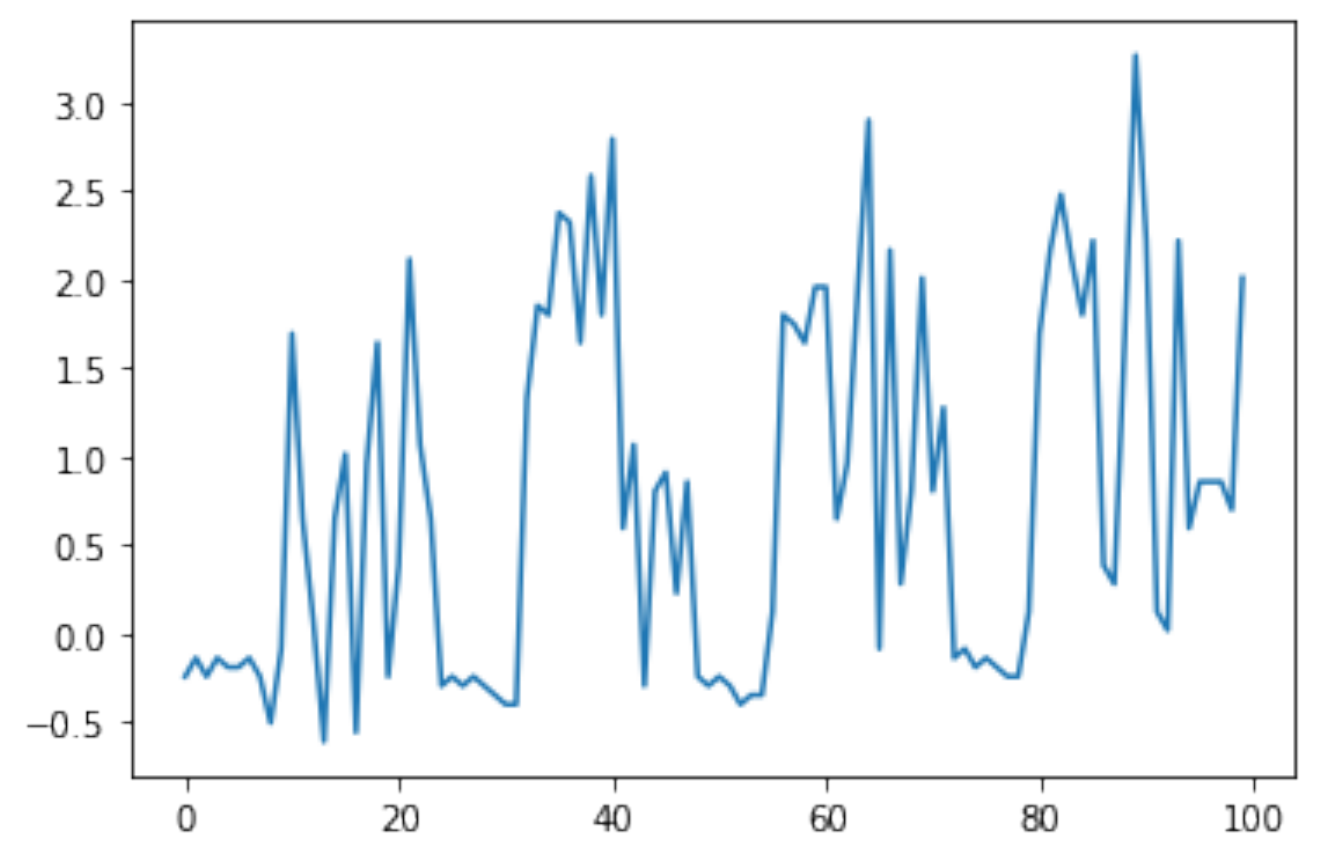}\label{fig:exp:original}}
    \subfigure[Jittering]{\includegraphics[width=1.3in]{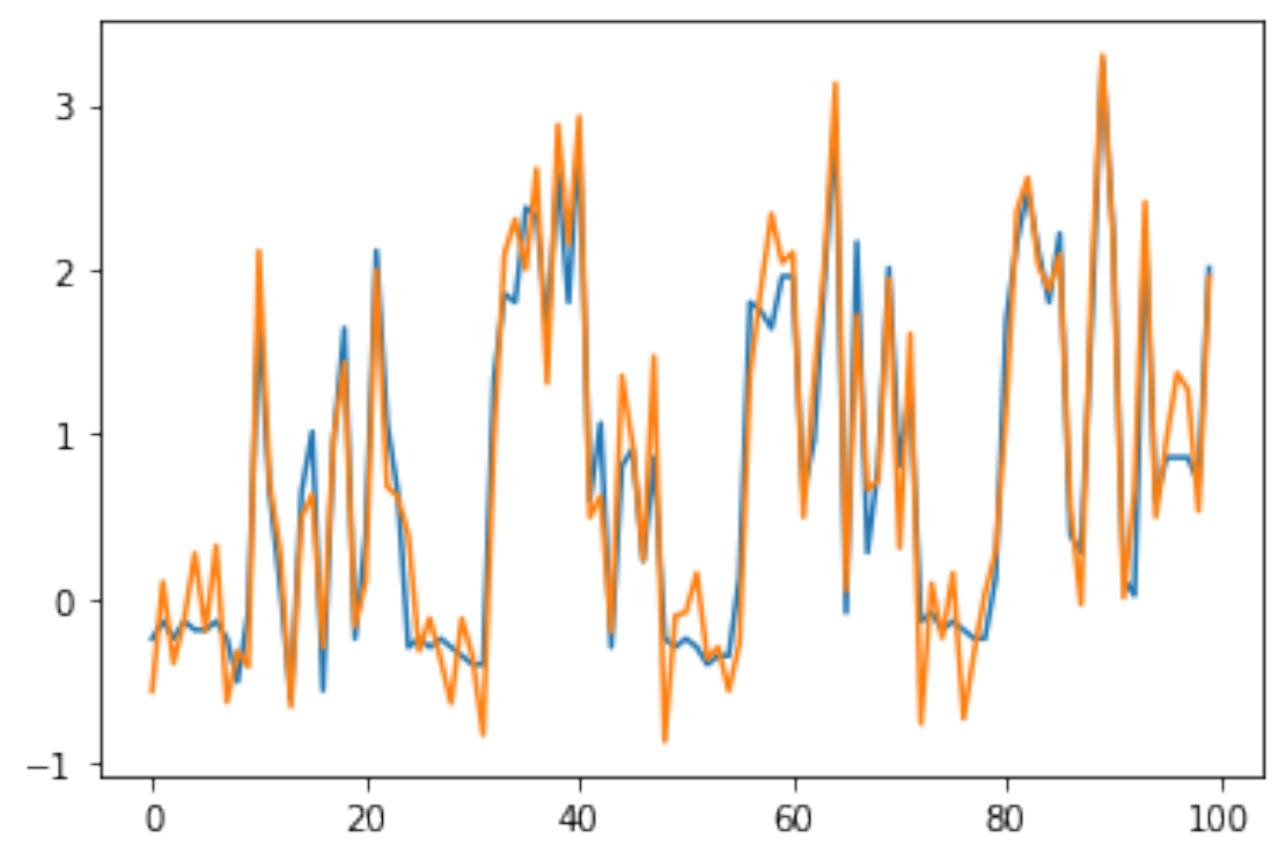}\label{fig:exp:jitter}}
    \subfigure[Scaling]{\includegraphics[width=1.3in]{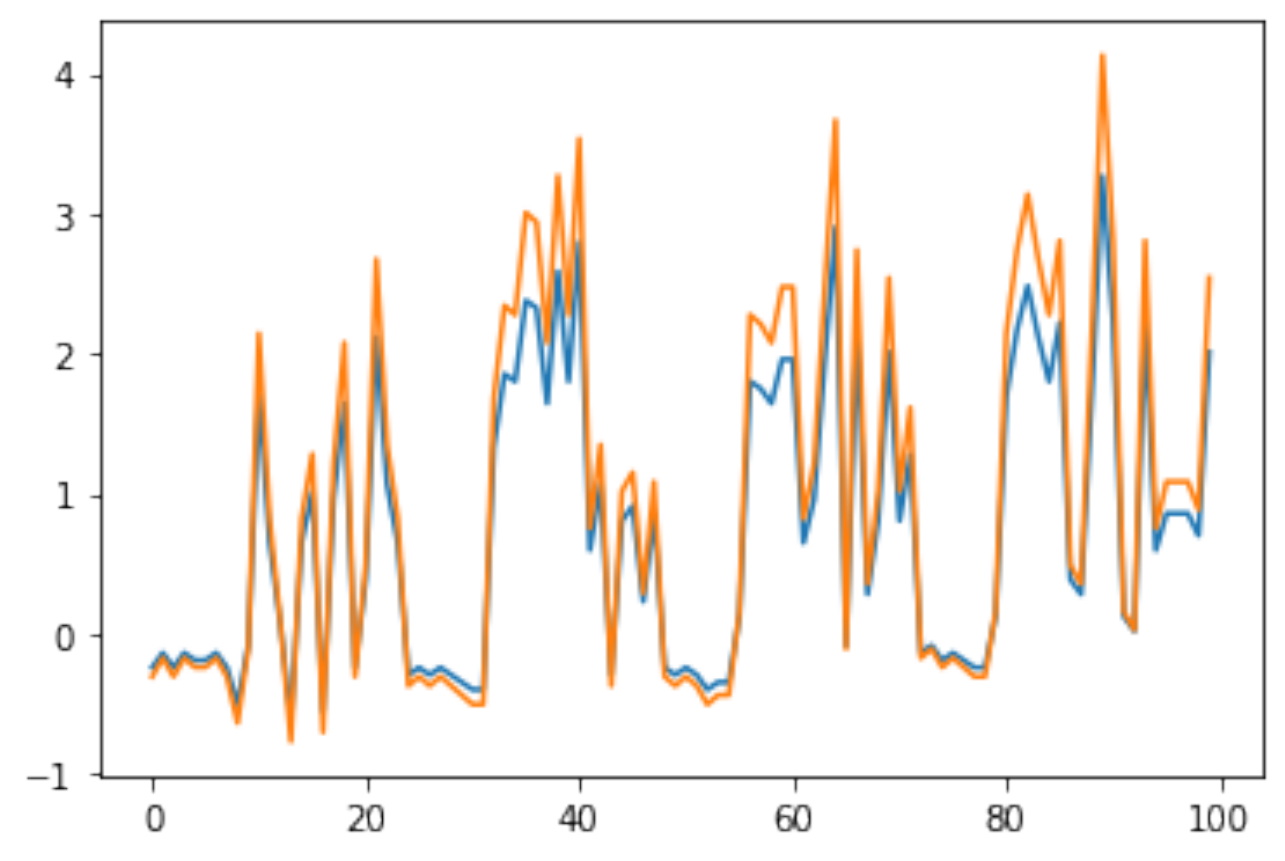}\label{fig:exp:scaling}}
    \subfigure[Cutout]{\includegraphics[width=1.3in]{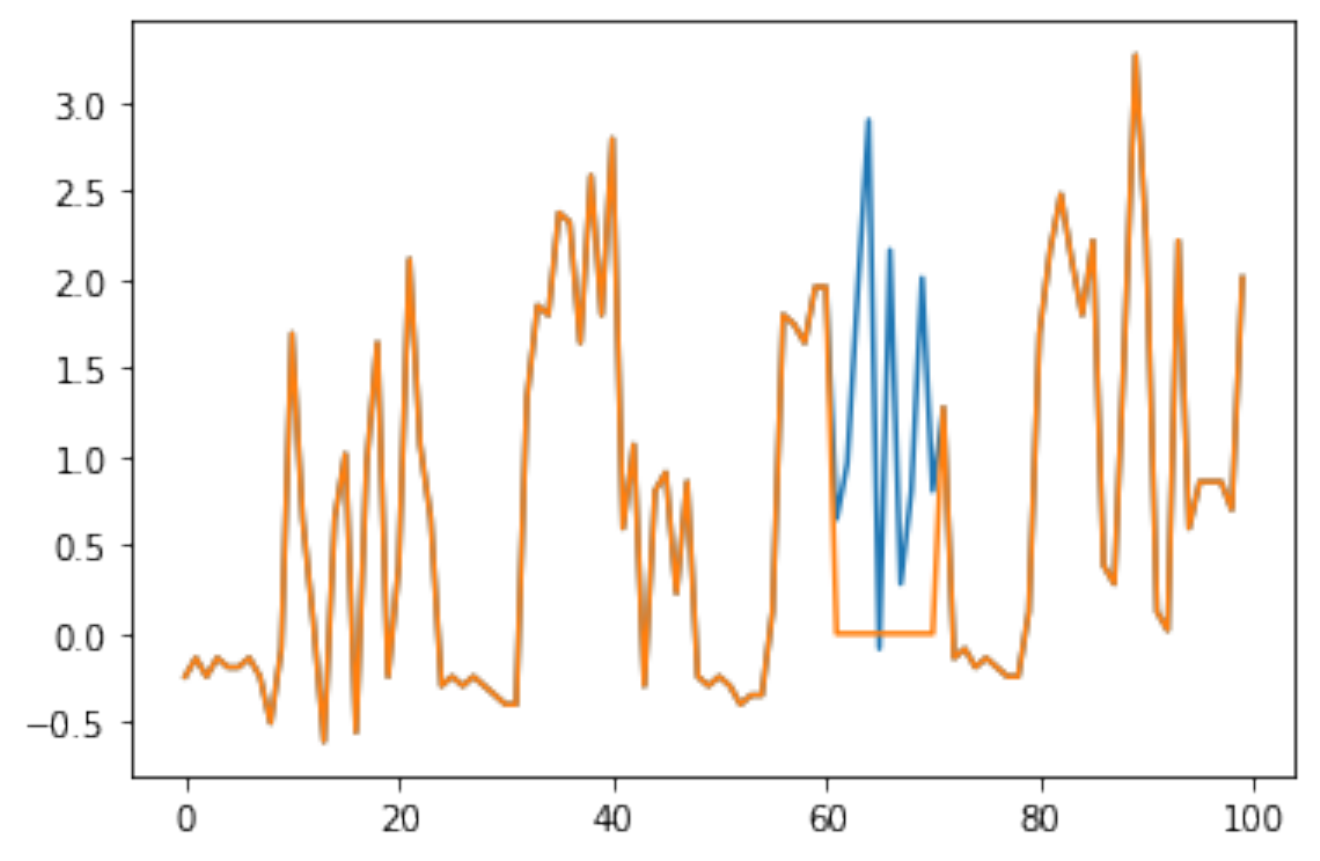}\label{fig:exp:cutout}}\\
    \subfigure[Time Warping]{\includegraphics[width=1.3in]{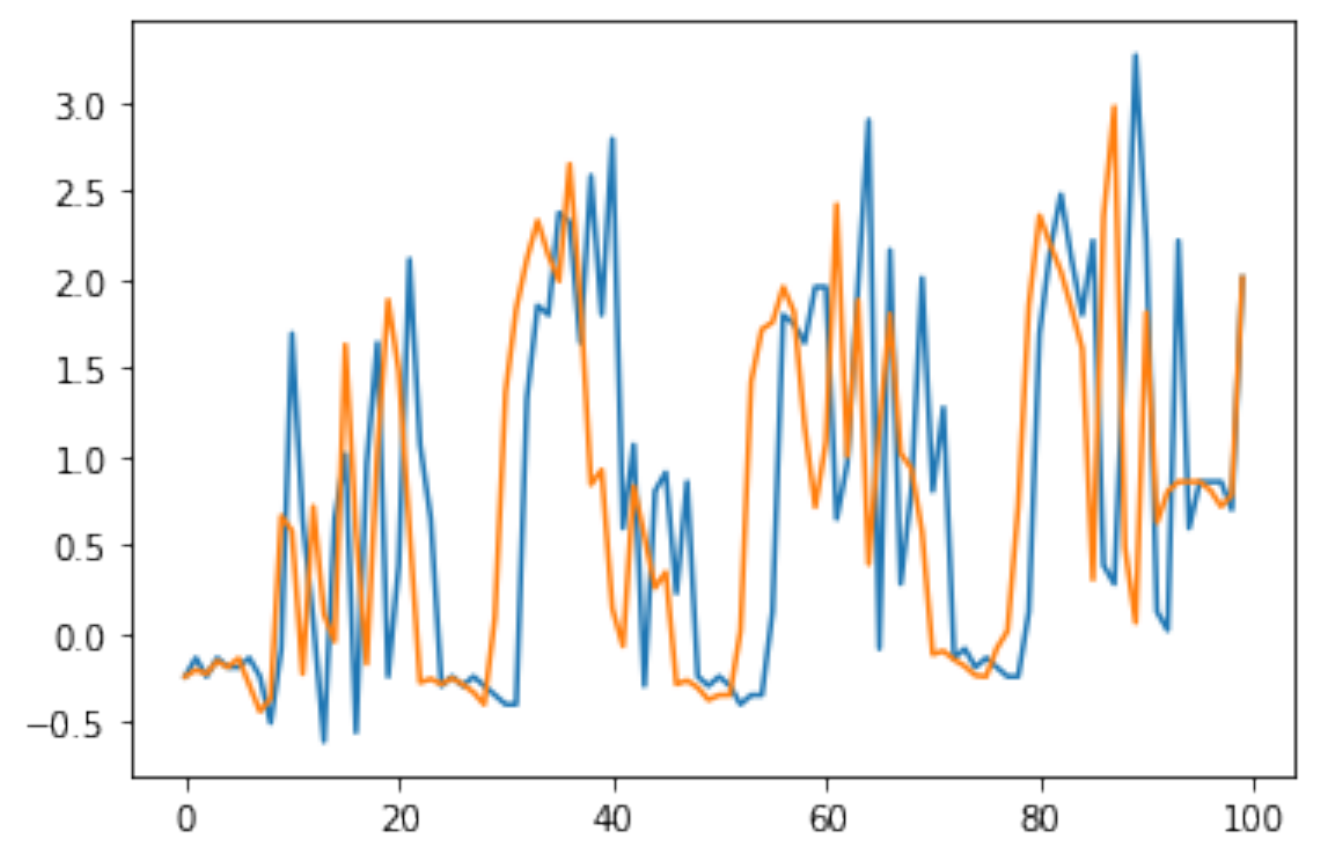}\label{fig:exp:timewarp}}
    \subfigure[Window Slicing]{\includegraphics[width=1.3in]{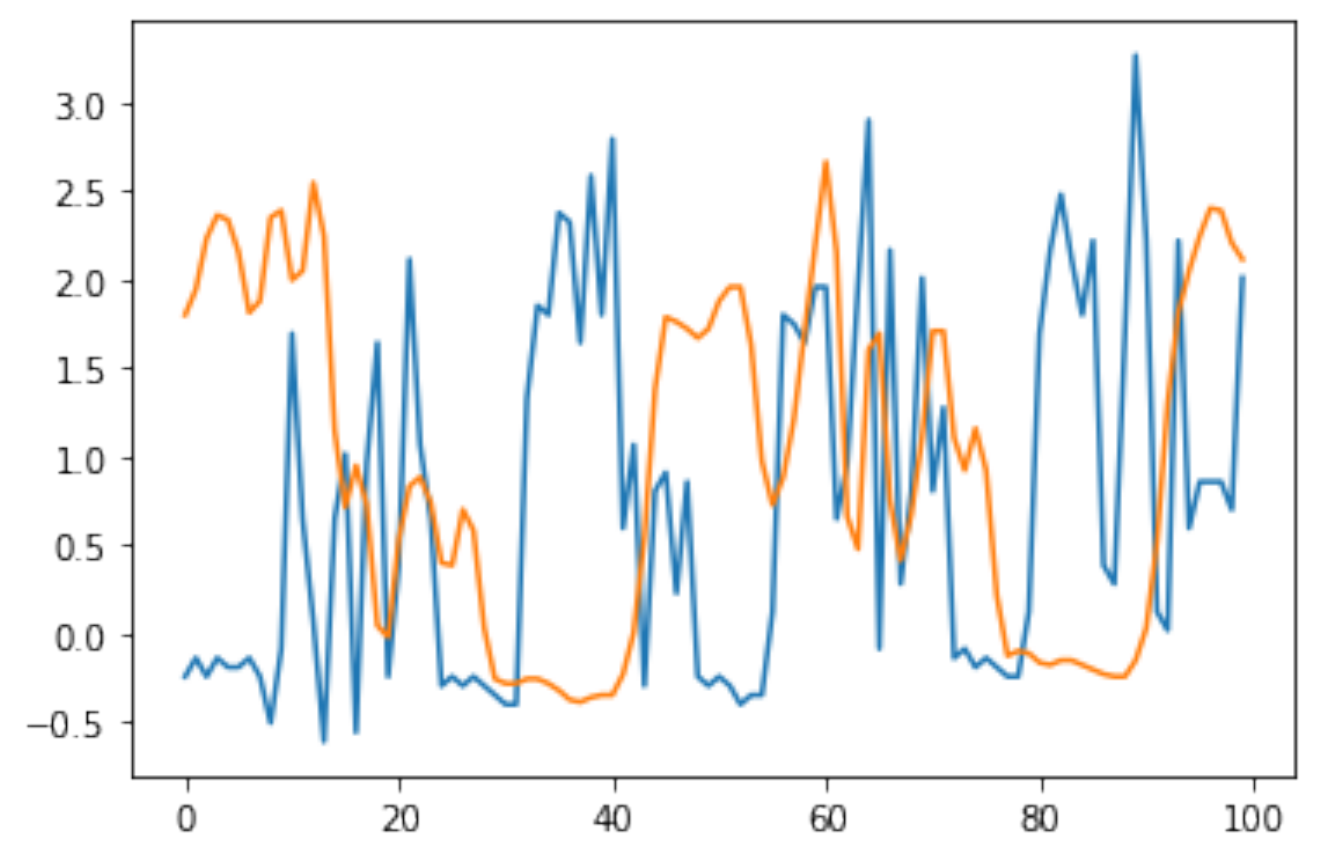}\label{fig:exp:windowslice}}
    \subfigure[Window Warping]{\includegraphics[width=1.3in]{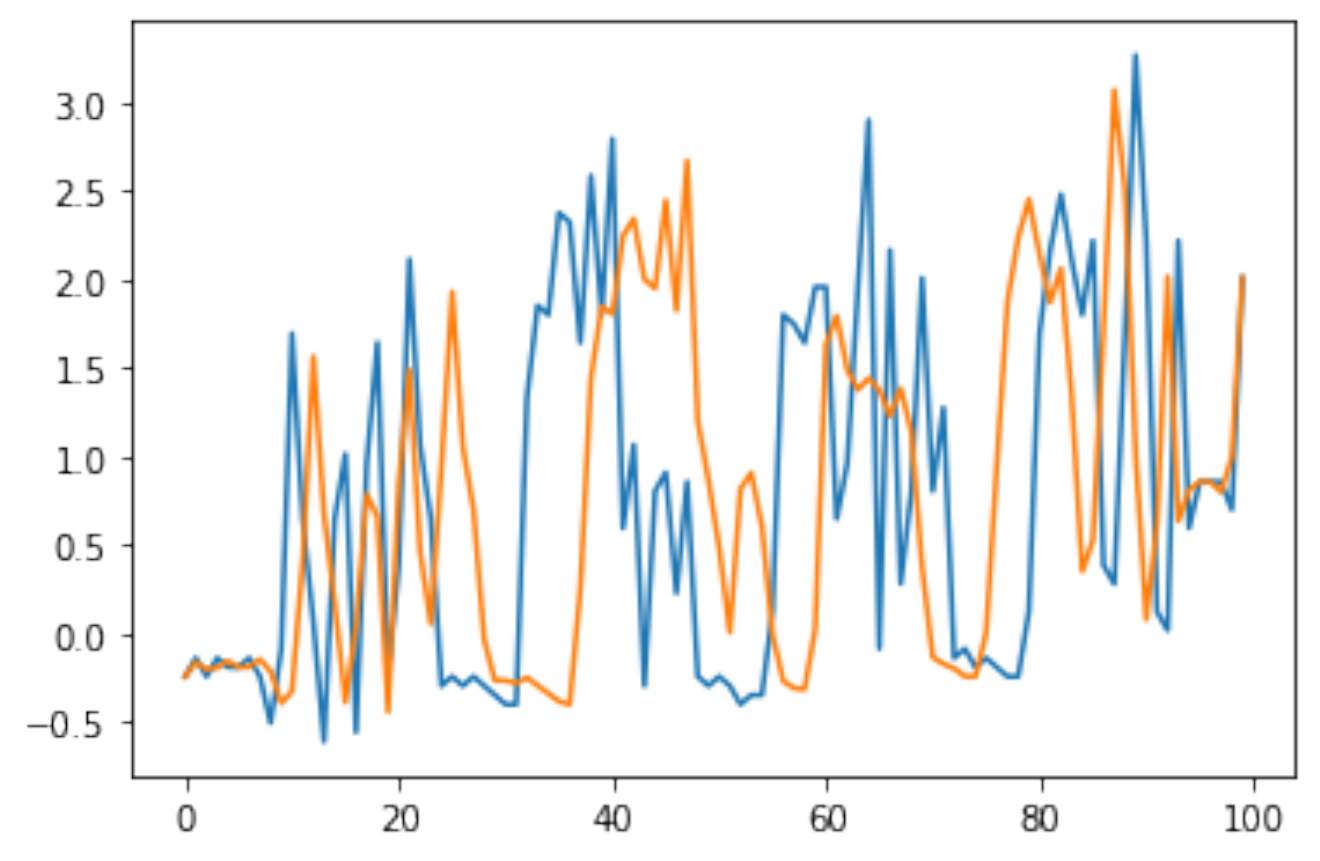}\label{fig:exp:windowwarp}}
    \subfigure[Subsequence]{\includegraphics[width=1.3in]{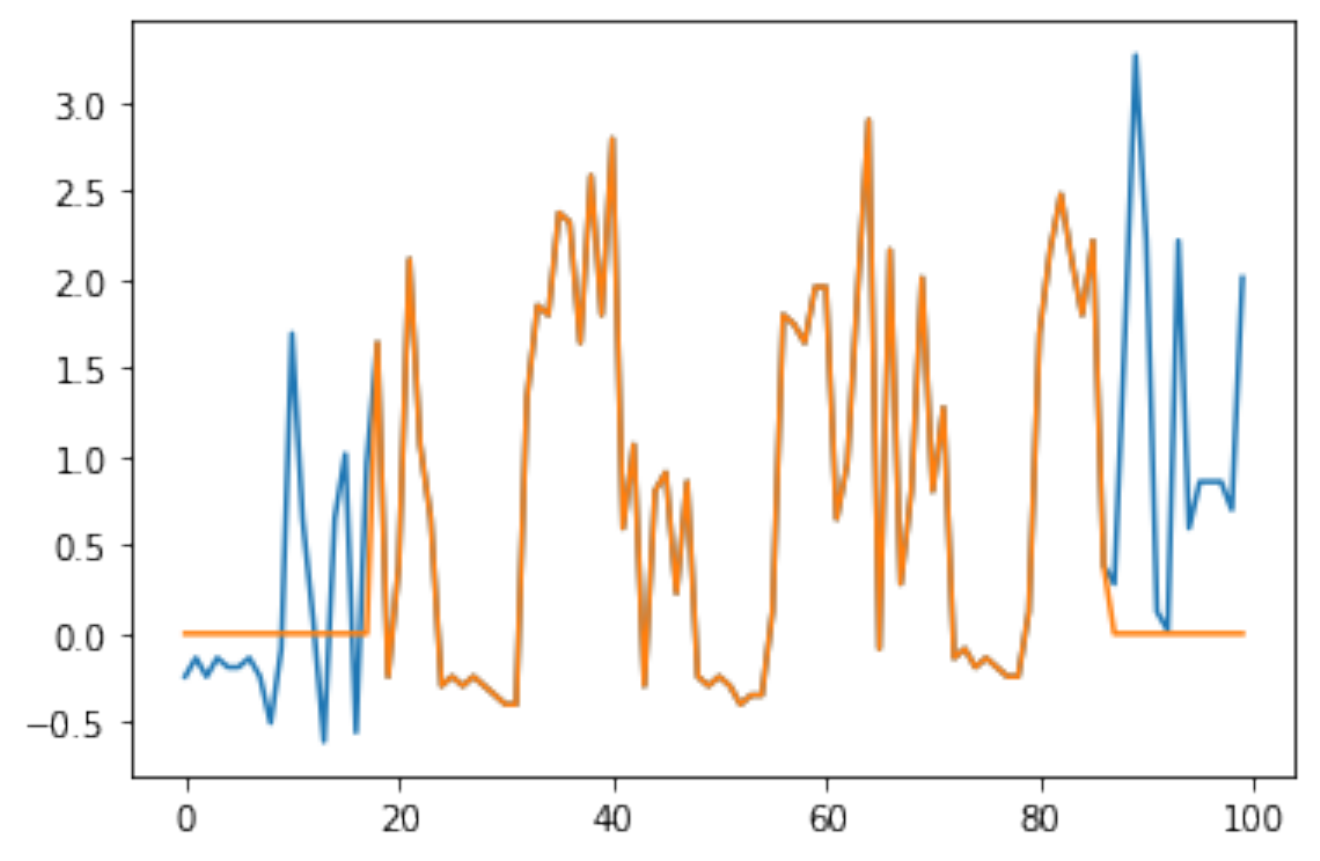}\label{fig:exp:Subsequence}}
    \caption{Examples of candidate augmentations on Electricity univariate dataset. Blue lines are the original time series data and orange ones are augmented instances.}
    \label{fig:augexample}
\end{figure}

With the first 100 time stamps in the univariate Electricity dataset as an example, we visualize the original time series and the augmented ones in Figure~\ref{fig:augexample}.

\subsection{Hardware and Implementations}
All experiments are conducted on a Linux machine with 4 NVIDIA GeForce RTX 2080 Ti GPUs, each with 11GB memory. CUDA version is 10.1 and Driver Version is 418.56.  Our method InfoTS is implemented with Python 3.7.7 and Pytorch 1.7.1.

\subsection{Hyperparameters}
We train and evaluate our methods with the following hyperparameters and configurations. 
\begin{itemize}
    \item Optimizer: Adam optimizer~\citep{kingma2014adam} with learning rate and decay rates setting to 0.001 and (0.9,0.999), respectively.
    \item SVM: scikit-learn implementation~\citep{scikit-learn} with penalty $C \in \{10^i | i\in [-4,4] \cup {\infty}\}$~\cite{franceschi2019unsupervised}.
    \item Encoder architecture: We follow~\citep{yue2021learning} to design the encoder. Specifically, the output dimension of the linear projection layer is set to 64, the same for the number of channels in the following dilated CNN module. In the CNN module, GELU~\citep{hendrycks2016gaussian} is adopted as the activation function, and the kernel size is set to 3. The dilation is set to $2^i$ in the $i$-the block.
    \item Classifier architecture: a fully connected layer that maps the representations to the label is adopted.
    \item Trade-off hyperparameters: $\beta$ in Eq.~(\ref{eq:criteria}) and $\alpha$ in Eq.~(\ref{eq:contrastive}) are searched in $[0.1,0.5,1.0,5,10]$. Parameter sensitivity studies are shown in Section~\ref{sec:exp:parameter}.
    \item Temperature in binary concrete distribution:  we follow the practice in~\citep{jang2016categorical} to adopt the strategy by starting the training with a high temperature 2.0, and anneal to a small value 0.1, with a guided schedule.
\end{itemize}

\section{More Experimental Results}
\label{sec:app:exp}

\begin{table*}
  \centering
    \caption{Multivariate time series forecasting results.}
  \label{tab:forecast-multivar}
  \scalebox{0.7}{
  \begin{tabular}{lccccccccccccccc}
  \toprule
         &  & \multicolumn{2}{c}{InfoTS} &
                  \multicolumn{2}{c}{TS2Vec} &
         \multicolumn{2}{c}{Informer} &
         \multicolumn{2}{c}{StemGNN} &
         \multicolumn{2}{c}{TCN} &
         \multicolumn{2}{c}{LogTrans} &
         \multicolumn{2}{c}{LSTnet} \\
    \cmidrule(r){3-4} \cmidrule(r){5-6} \cmidrule(r){7-8} \cmidrule(r){9-10} \cmidrule(r){11-12} \cmidrule(r){13-14}  \cmidrule(r){15-16}
    Dataset & $L_y$ & MSE & MAE & MSE & MAE & MSE & MAE & MSE & MAE & MSE & MAE & MSE & MAE & MSE & MAE \\
    \midrule
    \multirow{5}*{ETTh$_1$}
    & 24   & \textbf{0.564} & \textbf{0.520}     & 0.599 & 0.534 & 0.577 & 0.549 & 0.614 & 0.571 & 0.767 & 0.612 & 0.686 & 0.604 & 1.293 & 0.901 \\
    & 48 & \textbf{0.607} & \textbf{0.553} & 0.629 & 0.555 & 0.685 & 0.625 & 0.748 & 0.618 & 0.713 & 0.617 & 0.766 & 0.757 & 1.456 & 0.960 \\
    & 168 & 0.746 & 0.638 & 0.755 & 0.636 & 0.931 & 0.752 & \textbf{0.663} & \textbf{0.608} & 0.995 & 0.738 & 1.002 & 0.846 & 1.997 & 1.214 \\
    & 336 & \textbf{0.904} & 0.722 & 0.907 & \textbf{0.717} & 1.128 & 0.873 & 0.927 & 0.730 & 1.175 & 0.800 & 1.362 & 0.952 & 2.655 & 1.369 \\
    & 720& 1.098 & 0.811  & \textbf{1.048} & \textbf{0.790} & 1.215 & 0.896 & --\tnote{*} & -- & 1.453 & 1.311 & 1.397 & 1.291 & 2.143 & 1.380 \\
    \midrule
    \multirow{5}*{ETTh$_2$}
    & 24 & \textbf{0.383} & 0.462 & 0.398 & \textbf{0.461} & 0.720 & 0.665 & 1.292 & 0.883 & 1.365 & 0.888 & 0.828 & 0.750 & 2.742 & 1.457 \\
    & 48 & \textbf{0.567} & 0.582 & 0.578 & \textbf{0.573} & 1.457 & 1.001 & 1.099 & 0.847 & 1.395 & 0.960 & 1.806 & 1.034 & 3.567 & 1.687 \\
    & 168  & \textbf{1.789} & \textbf{1.048} & 1.901 & 1.065 & 3.489 & 1.515 & 2.282 & 1.228 & 3.166 & 1.407 & 4.070 & 1.681 & 3.242 & 2.513 \\
    & 336 & \textbf{2.120} & \textbf{1.161} & 2.304 & 1.215 & 2.723 & 1.340 & 3.086 & 1.351 & 3.256 & 1.481 & 3.875 & 1.763 & 2.544 & 2.591 \\
    & 720 & \textbf{2.511} & \textbf{1.316} &2.650 & 1.373 & 3.467 & 1.473 & -- & -- & 3.690 & 1.588 & 3.913 & 1.552 & 4.625 & 3.709 \\
    
    \midrule
 
    \multirow{5}*{ETTm$_1$}
    & 24 & 0.391 & 0.408 & 0.443 & 0.436 & \textbf{0.323}   & \textbf{0.369} & 0.620 & 0.570 & 0.324 & 0.374                   & 0.419 & 0.412 & 1.968 & 1.170 \\
    & 48 & 0.503 & 0.475 & 0.582 & 0.515 & 0.494            &       0.503    & 0.744 & 0.628 & \textbf{0.477} & \textbf{0.450} & 0.507 & 0.583 & 1.999 & 1.215 \\
    & 96 &\textbf{ 0.537} & \textbf{0.503} & 0.622 & 0.549 & 0.678 & 0.614 & 0.709 & 0.624 & 0.636 & 0.602 & 0.768 & 0.792 & 2.762 & 1.542 \\
    & 288  & \textbf{0.653} &\textbf{0.579} & 0.709 & 0.609 & 1.056 & 0.786 & 0.843 & 0.683 & 1.270 & 1.351 & 1.462 & 1.320 & 1.257 & 2.076 \\
    & 672 & \textbf{0.757} &\textbf{ 0.642} & 0.786 & 0.655 & 1.192 & 0.926 & -- & -- & 1.381 & 1.467 & 1.669 & 1.461 & 1.917 & 2.941 \\

    \midrule

    \multirow{5}*{Electricity}
    & 24 & \textbf{0.255} & \textbf{0.350} & 0.287 & 0.374 & 0.312 & 0.387 & 0.439 & 0.388 & 0.305 & 0.384 & 0.297 & 0.374 & 0.356 & 0.419 \\
    & 48 & \textbf{0.279} & \textbf{0.368} & 0.307 & 0.388 & 0.392 & 0.431 & 0.413 & 0.455 & 0.317 & 0.392 & 0.316 & 0.389 & 0.429 & 0.456 \\
    & 168 & \textbf{0.302} & \textbf{0.385} & 0.332 & 0.407 & 0.515 & 0.509 & 0.506 & 0.518 & 0.358 & 0.423 & 0.426 & 0.466 & 0.372 & 0.425 \\
    & 336 & \textbf{0.320} & \textbf{0.399} & 0.349 & 0.420 & 0.759 & 0.625 & 0.647 & 0.596 & 0.349 & 0.416 & 0.365 & 0.417 & 0.352 & 0.409 \\
        \midrule
    \multicolumn{2}{l}{Avg.}& \textbf{0.805} & \textbf{0.627} & 0.852 & 0.645 & 1.164 & 0.781 & 0.977 & 0.706 & 1.243 & 0.854 & 1.402  & 1.032 & 1.836 & 1.374 \\
    \bottomrule
  \end{tabular}
 }
\end{table*}

\subsection{Parameter Sensitivity Studies}
\label{sec:exp:parameter}
\begin{figure}[h]
    \centering
    \subfigure[$\alpha$ in Eq.~(\ref{eq:contrastive})]{\includegraphics[width=2.0in]{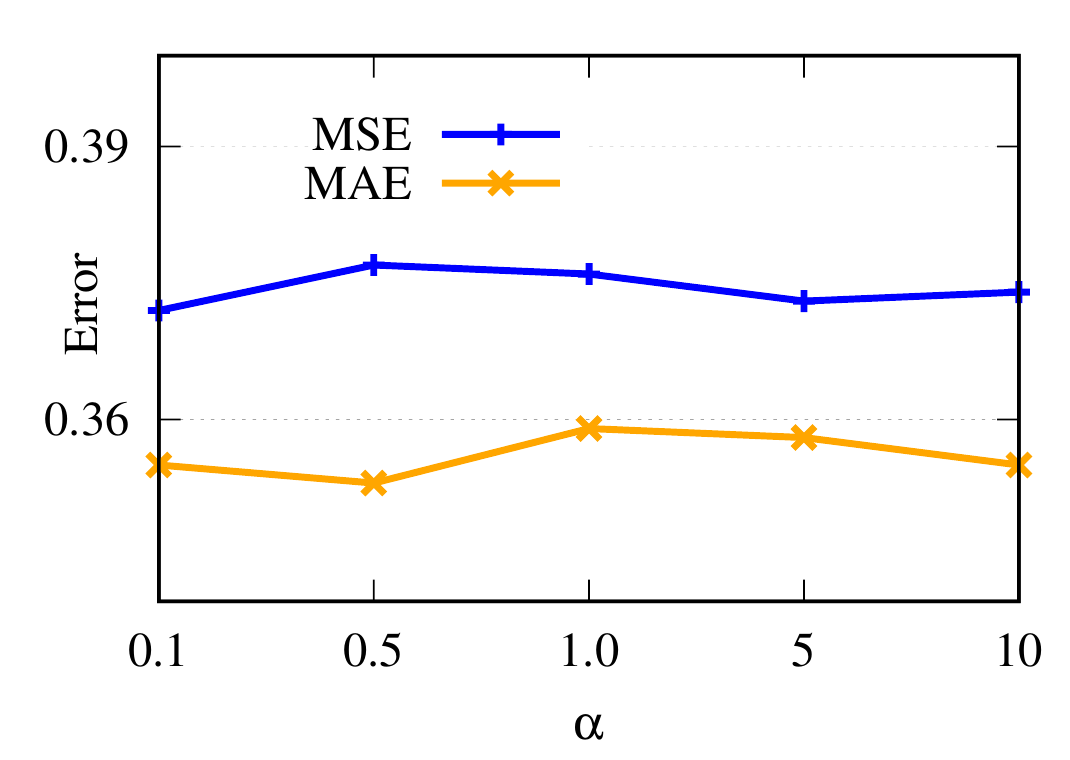}\label{fig:para:alpha}}
    \subfigure[$\beta$ in Eq.~(\ref{eq:criteria})]{\includegraphics[width=2.0in]{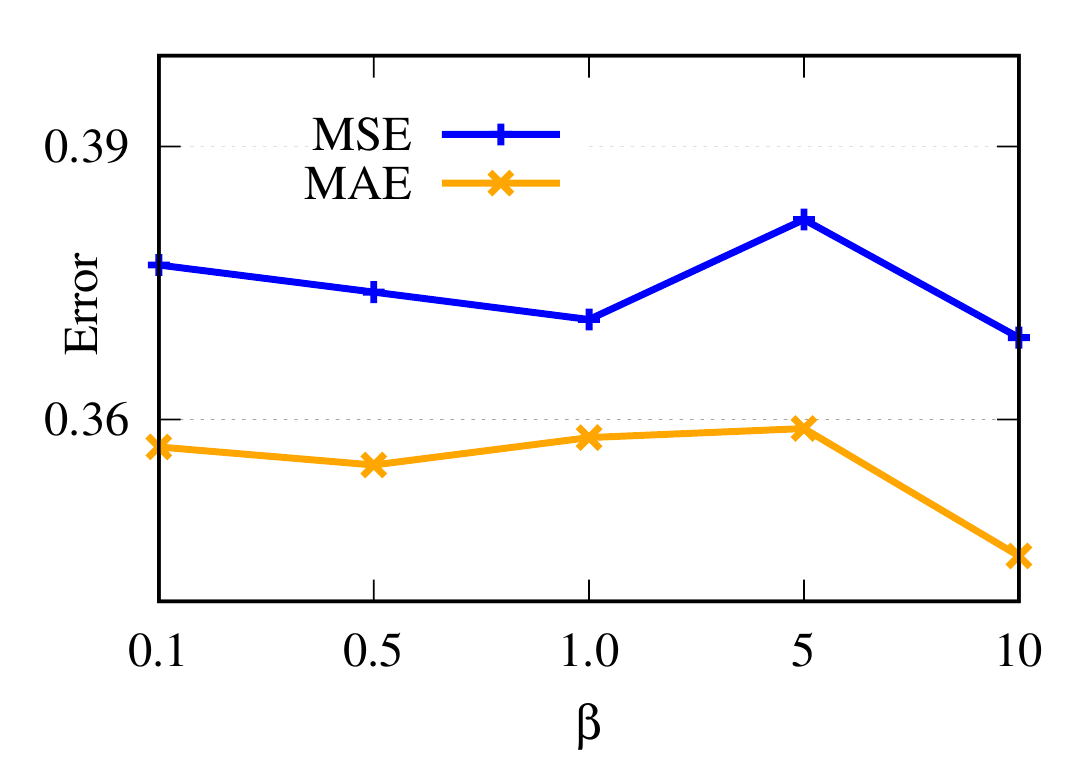}\label{fig:criteria:beta}}
    \caption{Parameter sensitivity studies.}
    \label{fig:exp:parameter}
\end{figure}

\normalsize
In this part, we adopt the electricity dataset to analyze the effects of two important hyper-parameters in our method InfoTS. The hyperparameter $\alpha$ in Eq.~(\ref{eq:contrastive}) controls the trade-off between local and global contrastive losses when training the encoder. $\beta$ in Eq.~(\ref{eq:criteria}) achieves the balance between high variety and high fidelity when training the meta-learner network. We tune these parameters in range $[0.1,0.5,1.0,5,10]$ and show the results in Figure~\ref{fig:exp:parameter}.  These figures show that our method achieves high performance with a wide range of selections, demonstrating the robustness of the proposed method. In general, setting trade-off parameters to 0.5 or 1 achieves good performance.

\subsection{Case Study on Signal Detection}
\begin{figure}[h]
    \centering
    \includegraphics[width=3in]{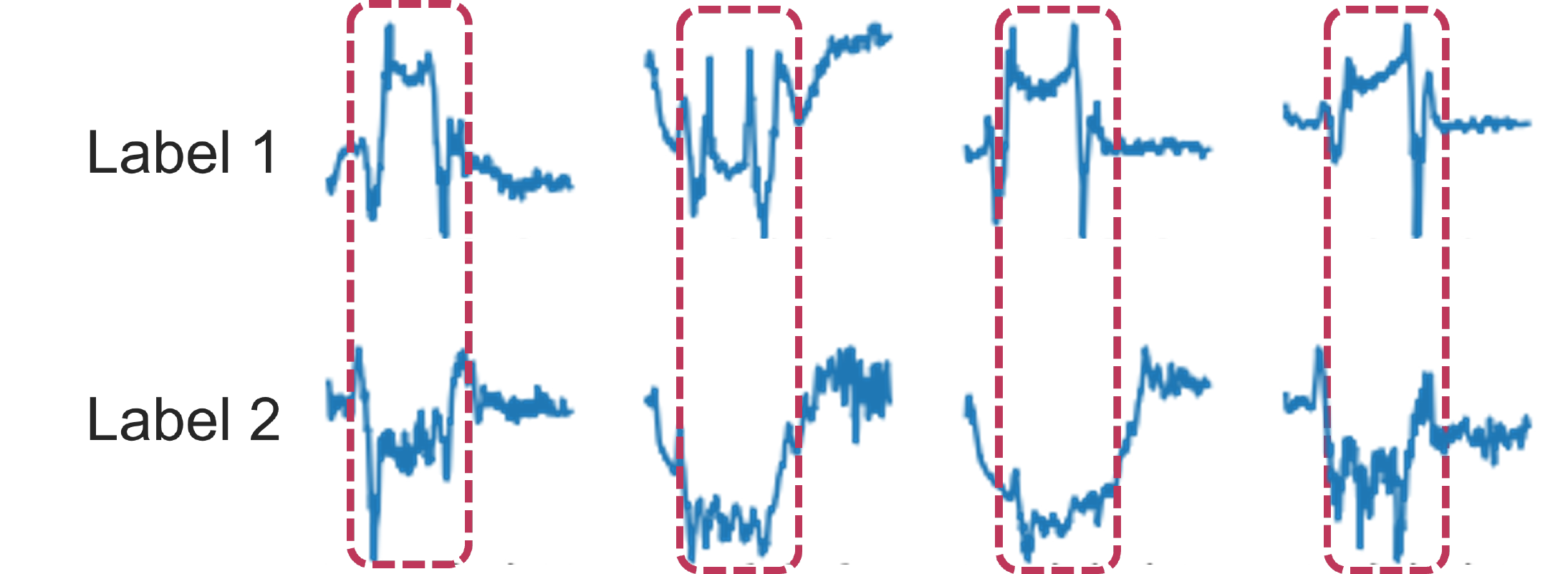}
    \caption{The informative signals locate in the middle periods of time series in the CricketX dataset.}
    \label{fig:signal}
\end{figure}

In this part, we show the potential usage of InfoTS to detect the informative signals in the time series. We adopt the CricketX dataset as an example for the case study. Subsequence augmentations on 0-100, 100-200, and 200-300 periods are adopted as candidate transformations. We observe that the one operated on 100-200 period has high fidelity and variety, leading to better accuracy performance, which is consistent with the visualization results in Figure~\ref{fig:signal}.

\subsection{Updating Process of InfoTS}
\label{sec:app:updating}

\begin{figure}
    \centering
    \includegraphics[width=3in]{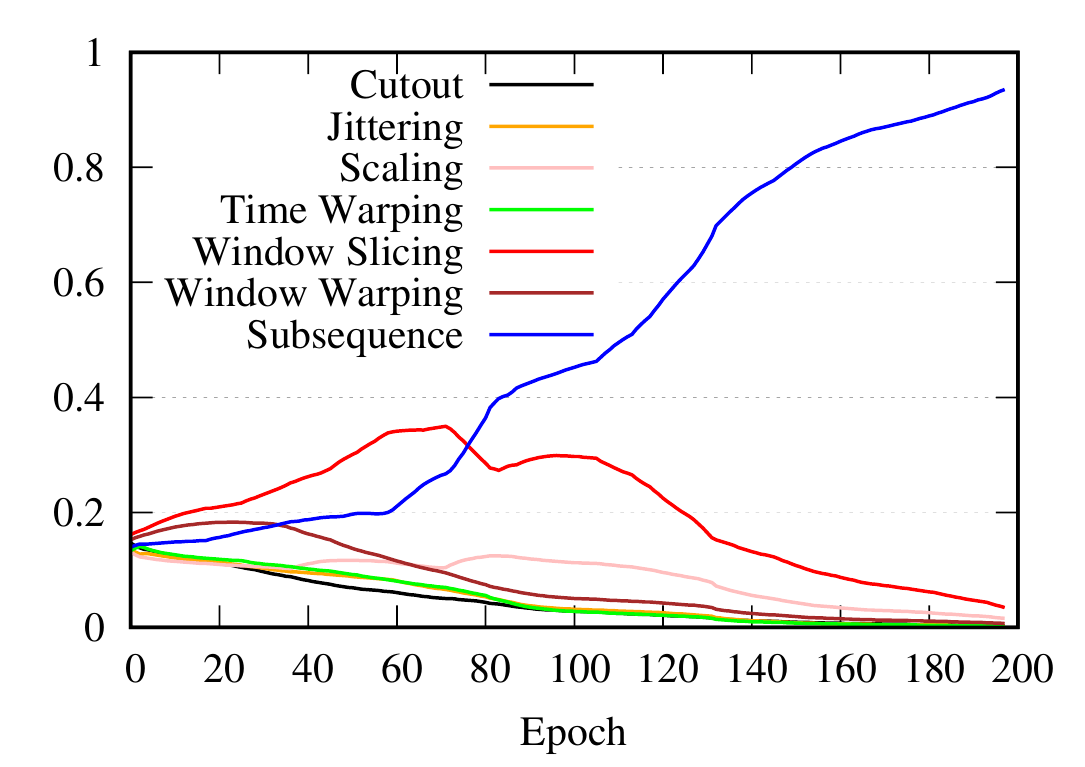}
    \caption{Weight updating process of meta-learner network in InfoTS.}
    \label{fig:weight}
\end{figure}

\begin{table*}[h]
  \centering
    \caption{Effectiveness of each candidate data transformation on Electricity.}
  \label{tab:eachaugmentation}
  \scalebox{0.67}{
  \begin{tabular}{lcccccccccccccccc}
  \toprule
        & \multicolumn{2}{c}{InfoTS} &
                  \multicolumn{2}{c}{Cutout} &
         \multicolumn{2}{c}{Jittering} &
         \multicolumn{2}{c}{Scaling} &
         \multicolumn{2}{c}{Time Warp} &
         \multicolumn{2}{c}{Window Slice} &
         \multicolumn{2}{c}{Window Warp} &
         \multicolumn{2}{c}{Subsequence} \\
    \cmidrule(r){2-3} \cmidrule(r){4-5} \cmidrule(r){6-7} \cmidrule(r){8-9} \cmidrule(r){10-11} \cmidrule(r){12-13}  \cmidrule(r){14-15} \cmidrule(r){16-17}
     $L_y$ & MSE & MAE & MSE & MAE & MSE & MAE & MSE & MAE & MSE & MAE & MSE & MAE & MSE & MAE & MSE & MAE\\
     \midrule
    24 & \textbf{0.245}  & \textbf{0.269} & 0.254 & 0.277 & 0.251 & 0.275  &0.252  & 0.273 & 0.251&  0.274  & 0.258 & 0.280 & 0.253 &0.277 &0.248 & 0.273\\
    48 & \textbf{0.294} & \textbf{0.301} & 0.304 & 0.309 & 0.297 & 0.302  &0.302  & 0.307 & 0.305 & 0.309  & 0.310 & 0.314 & 0.307 &0.310 &0.295 & \textbf{0.301}\\
    168 & \textbf{0.402} & \textbf{0.367} & 0.412 & 0.381 & 0.403 & 0.373 &0.407  & 0.377 & 0.415 & 0.382  & 0.415 & 0.381 & 0.416 &0.382 &0.405 &0.372\\
    336 & \textbf{0.533} & \textbf{0.453} & 0.555 & 0.465 &0.545 & 0.458  &0.552  & 0.461 &  0.555 &0.469  & 0.551 & 0.470 & 0.554 &0.466   &0.546 &0.456\\
         \midrule
    Avg. &\textbf{0.369} &	\textbf{0.348}	&0.381	&0.358	&0.374&	0.352&	0.377&	0.354&	0.381&	0.359&	0.383	&0.361&	0.383&	0.359&	0.374&	0.350 \\
    \bottomrule
  \end{tabular}
}
\end{table*}

\begin{table*}[h]
  \centering
    \caption{Effectiveness of each candidate data transformation on ETTh1.}
  \label{tab:eachaugmentation:etth1}
  \scalebox{0.67}{
  \begin{tabular}{lcccccccccccccccc}
  \toprule
        & \multicolumn{2}{c}{InfoTS} &
                  \multicolumn{2}{c}{Cutout} &
         \multicolumn{2}{c}{Jittering} &
         \multicolumn{2}{c}{Scaling} &
         \multicolumn{2}{c}{Time Warp} &
         \multicolumn{2}{c}{Window Slice} &
         \multicolumn{2}{c}{Window Warp} &
         \multicolumn{2}{c}{Subsequence} \\
    \cmidrule(r){2-3} \cmidrule(r){4-5} \cmidrule(r){6-7} \cmidrule(r){8-9} \cmidrule(r){10-11} \cmidrule(r){12-13}  \cmidrule(r){14-15} \cmidrule(r){16-17}
     $L_y$ & MSE & MAE & MSE & MAE & MSE & MAE & MSE & MAE & MSE & MAE & MSE & MAE & MSE & MAE & MSE & MAE\\
     \midrule
    24 & \textbf{0.039} & 0.149  & 0.045  & 0.158 & 0.045 & 0.160  &\textbf{0.039}  & \textbf{0.148} & 0.043 & 0.155  & 0.041 & 0.151 & 0.043 &0.156  &0.045 & 0.161\\
     48 &\textbf{0.056} & \textbf{0.179}  & 0.061  & 0.185 & 0.062 & 0.188  &0.060  & 0.185   & 0.061 & 0.186  & 0.064 & 0.190 & 0.063 &0.191  &0.063 & 0.188\\
    168 &\textbf{0.100} & \textbf{0.239}  & 0.110  & 0.251 & 0.115 & 0.261  &0.111  & 0.255   & 0.111 & 0.253  & 0.118 & 0.265 & 0.118 &0.265  &0.125 &0.271\\
     336 &\textbf{0.117} & \textbf{0.264} & 0.136  & 0.287 &0.127  & 0.278  &0.130  & 0.281   & 0.148 & 0.302  & 0.133 & 0.288 & 0.146 &0.303  &0.139 &0.291\\
    720 & \textbf{0.141} & \textbf{0.302} & 0.167  & 0.330 &0.143  & 0.304  &0.155  & 0.318   & 0.168 & 0.331  & 0.151 & 0.315 & 0.147 &0.308  &0.153 &0.314 \\
         \midrule
    Avg. &\textbf{0.091} &\textbf{0.227}	& 0.104	 & 0.242 &0.098  & 0.238  &0.099  & 0.237   & 0.106 & 0.246  & 0.101 & 0.242 & 0.103 &0.245  &0.105 &0.245 \\
    \bottomrule
  \end{tabular}
}
\end{table*}

To show that our InfoTS can adaptively detect the most effective augmentation based on the data distribution, we conduct more ablation studies to investigate comprehensively into the proposed model.  We compare performances of variants that each applies a single transformation to generate augmented instances in Table~\ref{tab:eachaugmentation}. From the table, we know that augmentation with subsequence benefits the most for the Electricity dataset.  We visualize the weight updating process of InfoTS in Figure~\ref{fig:weight}, with each line representing the normalized importance score of the corresponding transformation. The weight for subsequence increase with the epoch, showing that InfoTS tends to adopt subsequence as the optimal transformation. Consistency between accuracy performance and weight updating process demonstrates the effectiveness of InfoTS to adaptively select feasible transformations.  Besides, as shown in Table~\ref{tab:eachaugmentation}, InfoTS outperforms the variant that uses subsequence only. 

Note that although subsequence transformation works well for the Electricity dataset, it may generate uninformative augmented instances for other datasets. Table~\ref{tab:eachaugmentation:etth1} shows the performances of each single transformation in the ETTh1 dataset. The subsequence transformation is no more effective than other candidate transformations. Besides, guided by the information-aware criteria, our method InfoTS can still outperform other variants.  This comparison shows that the meta-learner network learns to consider the combinations, which is better than any (single) candidate augmentation. 

\subsection{Evaluation of The Criteria with Time series Classification}
\begin{figure}
    \centering
    \subfigure[Supervised Setting]{\includegraphics[width=1.2in]{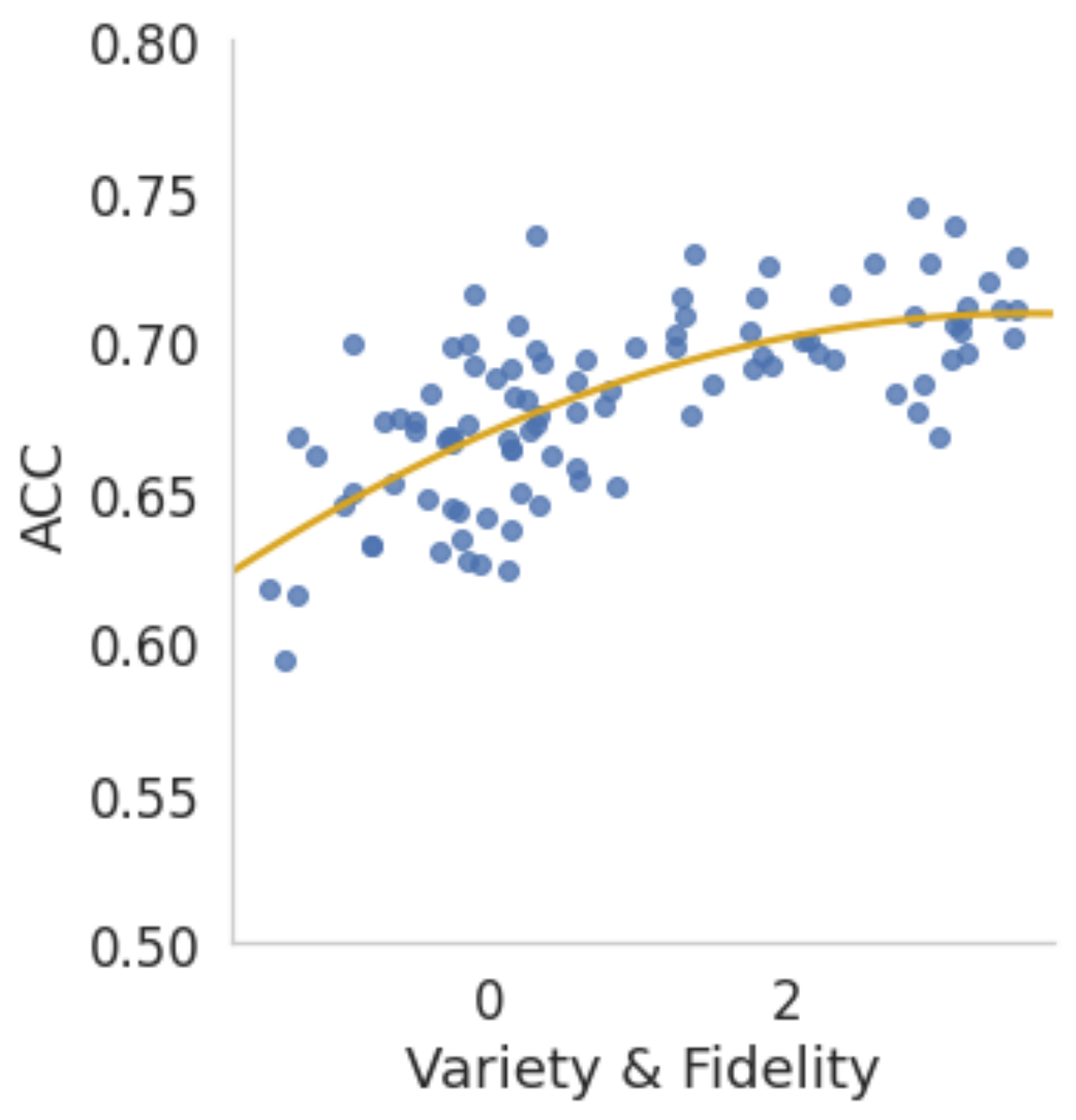}\label{fig:criteria:sup}}\quad\quad\quad\quad
    \subfigure[Unsupervised Setting]{\includegraphics[width=1.2in]{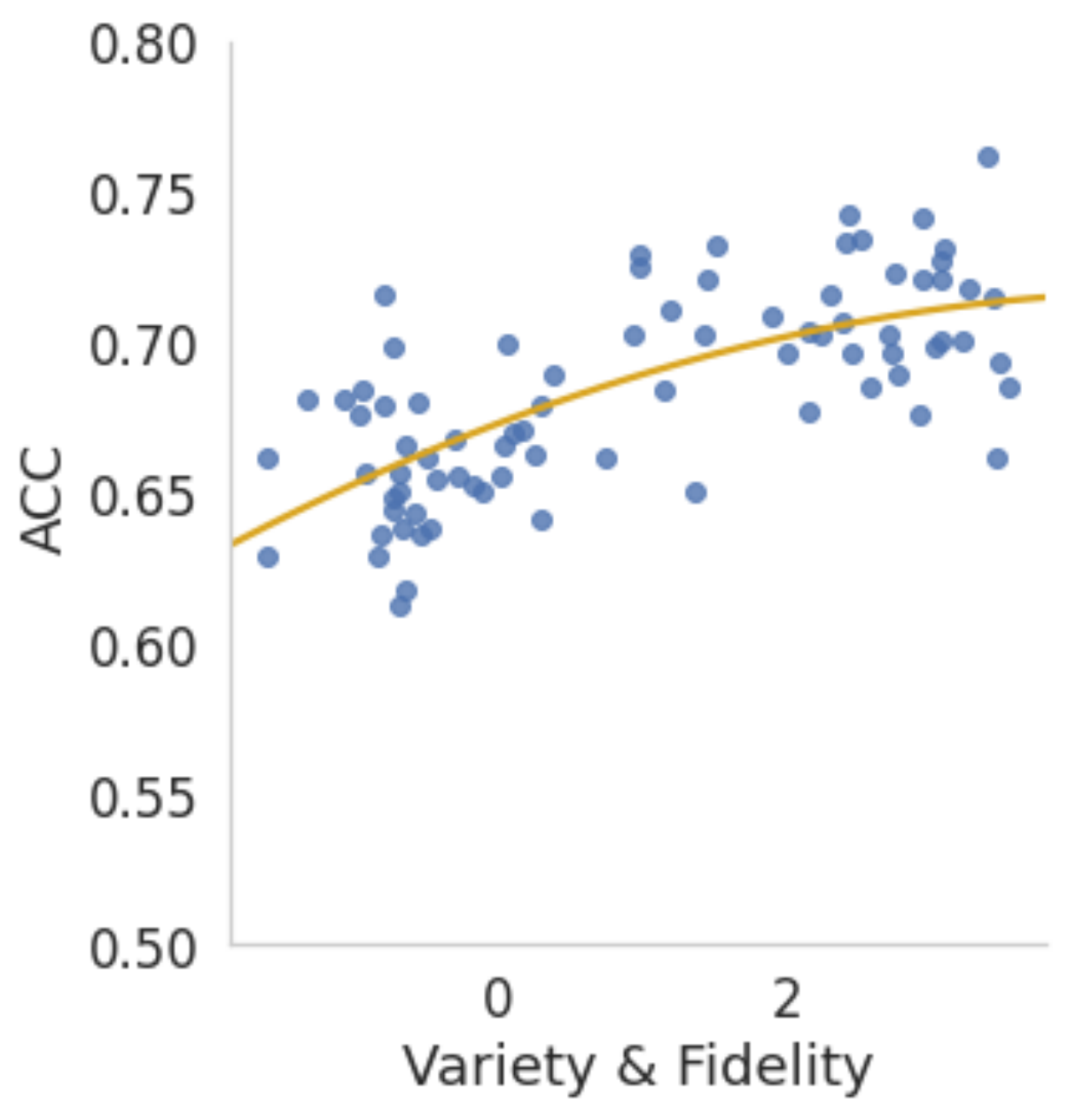}\label{fig:criteria:unsup}}\vspace{-0.2cm}
    \caption{Evaluation of the criteria.}
    \label{fig:app:criteria}
\end{figure}

To empirically verify the effectiveness of the proposed criteria with Time series Classification in both supervised and unsupervised setting, we adopt the dataset CricketY from the UCR archive~\citep{dau2019ucr,fan2020self}, and conduct augmentations with different configurations. For each configuration, we calculate the criteria score and the corresponding classification accuracy within the setting in Section~\ref{sec:exp:classifcation}. As shown in Figure~\ref{fig:app:criteria}, in general, accuracy performance is positively related to the proposed criteria in both supervised and unsupervised settings, the results are consistent with the conclusion drawn from forecasting performances.

\subsection{More Ablation Studies}
\label{sec:app:ablation}
To further check the effectiveness of the meta-learner network on automatically selecting suitable augmentations, we adopt another state-of-the-art baseline, TS2Vec as the backbone~\cite{yue2021learning}. We denote this variant as TS2Vec+Infoadpative, where the contrastive loss in TS2Vec is adopted to train the encoder and the proposed information-aware criteria are used to train the meta-learner network. The performances of the original TS2Vec and TS2Vec+Infoadpative on the Electricity dataset are shown in Table~\ref{tab:app:ts2vecinfo}. The comparison shows that with information-aware adaptive augmentation, we can also consistently and significantly improve the performances of TS2vec.

\begin{table}[h]
  \centering
    \caption{Effectiveness of meta-learner network with TS2Vec as the backbone.}
  \label{tab:app:ts2vecinfo}
  \scalebox{0.8}{
  \begin{tabular}{lcccc}
  \toprule
        & \multicolumn{2}{c}{TS2vec} &
                  \multicolumn{2}{c}{TS2Vec+Infoadpative}\\
    \cmidrule(r){2-3} \cmidrule(r){4-5} 
     $L_y$ & MSE & MAE & MSE & MAE \\
     \midrule
    24 &0.260  &0.288 &0.250 &0.273  \\
    48 &0.319  &0.324 &0.298 &0.302  \\
    168 &0.427  &0.394 &0.411 &0.372  \\
    336 &0.565  &0.474 &0.561 &0.463  \\
     \midrule
    Avg.  &0.393  &0.370 &0.380 &0.352  \\
    \bottomrule
  \end{tabular}
}
\end{table}

\subsection{Full Results of Time Series Forecasting and Classification}
\label{sec:app:full}
The full results of multivariate time series forecasting are shown in Tabel~\ref{tab:forecast-multivar}. Results of StemGNN with $L_y=720$ are not available due to the out-of-memory error~\citep{yue2021learning}.
Full results of univariate time series classification on 128 UCR datasets are shown in Tabel~\ref{tab:app:full-ucr}. 
Results of T-Loss, TS-TCC, and TNC are not reported on several datasets because they are not able to deal with missing observations in time series data. These unavailable accuracy scores are dismissed when computing average accuracy and considered as 0 when calculating the average rank. Results of multivariate classification on 30 UEA datasets are listed in Tabel~\ref{tab:app:full-uea}. Computations of average accuracy scores and ranks follow the ones in 128 UCR datasets.

\newpage
\onecolumn 

\relsize{-1}{
\begin{longtable}{lcccccccc}
  \toprule
     & InfoTS$_s$ &     InfoTS &      TS2Vec &      T-Loss &     TNC&     TS-TCC &     TST&    DTW  \\
    \midrule
    \endhead
Adiac & \textbf{0.795} & 0.788 & 0.775 & 0.675 & 0.726 & 0.767 & 0.550 & 0.604\\
ArrowHead & \textbf{0.874} & \textbf{0.874} & 0.857 & 0.766 & 0.703 & 0.737 & 0.771 & 0.703\\
Beef & \textbf{0.900} & 0.833 & 0.767 & 0.667 & 0.733 & 0.600 & 0.500 & 0.633\\
BeetleFly & 0.950 & 0.950 & 0.900 & 0.800 & 0.850 & 0.800 & \textbf{1.000} & 0.700\\
BirdChicken & 0.850 & \textbf{0.900} & 0.800 & 0.850 & 0.750 & 0.650 & 0.650 & 0.750\\
Car & \textbf{0.900} & 0.883 & 0.883 & 0.833 & 0.683 & 0.583 & 0.550 & 0.733\\
CBF & \textbf{1.000} & 0.999 & \textbf{1.000} & 0.983 & 0.983 & 0.998 & 0.898 & 0.997\\
ChlorineConcentration & 0.825 & 0.822 & \textbf{0.832} & 0.749 & 0.760 & 0.753 & 0.562 & 0.648\\
CinCECGTorso & 0.896 & \textbf{0.928} & 0.827 & 0.713 & 0.669 & 0.671 & 0.508 & 0.651\\
Coffee & \textbf{1.000} & \textbf{1.000} & \textbf{1.000} & \textbf{1.000} & \textbf{1.000} & \textbf{1.000} & 0.821 & \textbf{1.000}\\
Computers & 0.720 & \textbf{0.748} & 0.660 & 0.664 & 0.684 & 0.704 & 0.696 & 0.700\\
CricketX & 0.780 & 0.774 & \textbf{0.805} & 0.713 & 0.623 & 0.731 & 0.385 & 0.754\\
CricketY & \textbf{0.774} & \textbf{0.774} & 0.769 & 0.728 & 0.597 & 0.718 & 0.467 & 0.744\\
CricketZ & \textbf{0.792} & 0.787 & \textbf{0.792} & 0.708 & 0.682 & 0.713 & 0.403 & 0.754\\
DiatomSizeReduction & \textbf{0.997} & \textbf{0.997} & 0.987 & 0.984 & 0.993 & 0.977 & 0.961 & 0.967\\
DistalPhalanxOutlineCorrect & \textbf{0.808} & 0.801 & 0.775 & 0.775 & 0.754 & 0.754 & 0.728 & 0.717\\
DistalPhalanxOutlineAgeGroup & 0.763 & 0.763 & 0.727 & 0.727 & 0.741 & 0.755 & 0.741 & \textbf{0.770}\\
DistalPhalanxTW & 0.720 & \textbf{0.727} & 0.698 & 0.676 & 0.669 & 0.676 & 0.568 & 0.590\\
Earthquakes & \textbf{0.821} & \textbf{0.821} & 0.748 & 0.748 & 0.748 & 0.748 & 0.748 & 0.719\\
ECG200 & \textbf{0.950} & 0.930 & 0.920 & 0.940 & 0.830 & 0.880 & 0.830 & 0.770\\
ECG5000 & \textbf{0.945} & \textbf{0.945} & 0.935 & 0.933 & 0.937 & 0.941 & 0.928 & 0.924\\
ECGFiveDays & \textbf{1.000} & \textbf{1.000} & \textbf{1.000} & \textbf{1.000} & 0.999 & 0.878 & 0.763 & 0.768\\
ElectricDevices & 0.691 & 0.702 & \textbf{0.721} & 0.707 & 0.700 & 0.686 & 0.676 & 0.602\\
FaceAll & \textbf{0.929} & \textbf{0.929} & 0.805 & 0.786 & 0.766 & 0.813 & 0.504 & 0.808\\
FaceFour & 0.864 & 0.818 & \textbf{0.932} & 0.920 & 0.659 & 0.773 & 0.511 & 0.830\\
FacesUCR & 0.917 & 0.913 & \textbf{0.930} & 0.884 & 0.789 & 0.863 & 0.543 & 0.905\\
FiftyWords & \textbf{0.809} & 0.793 & 0.774 & 0.732 & 0.653 & 0.653 & 0.525 & 0.690\\
Fish & \textbf{0.949} & 0.937 & 0.937 & 0.891 & 0.817 & 0.817 & 0.720 & 0.920\\
FordA & 0.925 & 0.915 & \textbf{0.948} & 0.928 & 0.902 & 0.930 & 0.568 & 0.555\\
FordB & 0.795 & 0.785 & 0.807 & 0.793 & 0.733 & \textbf{0.815} & 0.507 & 0.620\\
GunPoint & \textbf{1.000} & \textbf{1.000} & 0.987 & 0.980 & 0.967 & 0.993 & 0.827 & 0.907\\
Ham & \textbf{0.848} & 0.838 & 0.724 & 0.724 & 0.752 & 0.743 & 0.524 & 0.467\\
HandOutlines & \textbf{0.946} & \textbf{0.946} & 0.930 & 0.922 & 0.930 & 0.724 & 0.735 & 0.881\\
Haptics & 0.545 & \textbf{0.546} & 0.536 & 0.490 & 0.474 & 0.396 & 0.357 & 0.377\\
Herring & \textbf{0.703} & 0.656 & 0.641 & 0.594 & 0.594 & 0.594 & 0.594 & 0.531\\
InlineSkate & 0.420 & \textbf{0.424} & 0.415 & 0.371 & 0.378 & 0.347 & 0.287 & 0.384\\
InsectWingbeatSound & \textbf{0.664} & 0.639 & 0.630 & 0.597 & 0.549 & 0.415 & 0.266 & 0.355\\
ItalyPowerDemand & \textbf{0.971} & 0.966 & 0.961 & 0.954 & 0.928 & 0.955 & 0.845 & 0.950\\
LargeKitchenAppliances & 0.851 & 0.853 & \textbf{0.875} & 0.789 & 0.776 & 0.848 & 0.595 & 0.795\\
Lightning2 & \textbf{0.934} & \textbf{0.934} & 0.869 & 0.869 & 0.869 & 0.836 & 0.705 & 0.869\\
Lightning7 & 0.863 & \textbf{0.877} & 0.863 & 0.795 & 0.767 & 0.685 & 0.411 & 0.726\\
Mallat & 0.967 & \textbf{0.974} & 0.915 & 0.951 & 0.871 & 0.922 & 0.713 & 0.934\\
Meat & \textbf{0.967} & \textbf{0.967} & \textbf{0.967} & 0.950 & 0.917 & 0.883 & 0.900 & 0.933\\
MedicalImages & \textbf{0.920} & 0.820 & 0.793 & 0.750 & 0.754 & 0.747 & 0.632 & 0.737\\
MiddlePhalanxOutlineCorrect & \textbf{0.859} & \textbf{0.859} & 0.838 & 0.825 & 0.818 & 0.818 & 0.753 & 0.698\\
MiddlePhalanxOutlineAgeGroup & \textbf{0.662} & \textbf{0.662} & 0.636 & 0.656 & 0.643 & 0.630 & 0.617 & 0.500\\
MiddlePhalanxTW & \textbf{0.636} & 0.617 & 0.591 & 0.591 & 0.571 & 0.610 & 0.506 & 0.506\\
MoteStrain & \textbf{0.873} & \textbf{0.873} & 0.863 & 0.851 & 0.825 & 0.843 & 0.768 & 0.835\\
NonInvasiveFetalECGThorax1 & \textbf{0.941} & \textbf{0.941} & 0.930 & 0.878 & 0.898 & 0.898 & 0.471 & 0.790\\
NonInvasiveFetalECGThorax2 & 0.943 & \textbf{0.944} & 0.940 & 0.919 & 0.912 & 0.913 & 0.832 & 0.865\\
OliveOil & \textbf{0.933} & \textbf{0.933} & 0.900 & 0.867 & 0.833 & 0.800 & 0.800 & 0.833\\
OSULeaf & 0.760 & 0.760 & \textbf{0.876} & 0.760 & 0.723 & 0.723 & 0.545 & 0.591\\
PhalangesOutlinesCorrect & \textbf{0.826} & \textbf{0.826} & 0.823 & 0.784 & 0.787 & 0.804 & 0.773 & 0.728\\
Phoneme & 0.272 & 0.281 & \textbf{0.312} & 0.276 & 0.180 & 0.242 & 0.139 & 0.228\\
Plane & \textbf{1.000} & \textbf{1.000} & \textbf{1.000} & 0.990 & \textbf{1.000} & \textbf{1.000} & 0.933 & \textbf{1.000}\\
ProximalPhalanxOutlineCorrect & 0.924 & \textbf{0.927} & 0.900 & 0.859 & 0.866 & 0.873 & 0.770 & 0.784\\
ProximalPhalanxOutlineAgeGroup & \textbf{0.883} & \textbf{0.883} & 0.844 & 0.844 & 0.854 & 0.839 & 0.854 & 0.805\\
ProximalPhalanxTW & \textbf{0.849} & 0.844 & 0.824 & 0.771 & 0.810 & 0.800 & 0.780 & 0.761\\
RefrigerationDevices & \textbf{0.624} & \textbf{0.624} & 0.589 & 0.515 & 0.565 & 0.563 & 0.483 & 0.464\\
ScreenType & \textbf{0.510} & 0.493 & 0.411 & 0.416 & 0.509 & 0.419 & 0.419 & 0.397\\
ShapeletSim & 0.856 & 0.856 & \textbf{1.000} & 0.672 & 0.589 & 0.683 & 0.489 & 0.650\\
ShapesAll & 0.855 & 0.852 & \textbf{0.905} & 0.848 & 0.788 & 0.773 & 0.733 & 0.768\\
SmallKitchenAppliances & \textbf{0.773} & \textbf{0.773} & 0.733 & 0.677 & 0.725 & 0.691 & 0.592 & 0.643\\
SonyAIBORobotSurface1 & 0.921 & \textbf{0.927} & 0.903 & 0.902 & 0.804 & 0.899 & 0.724 & 0.725\\
SonyAIBORobotSurface2 & \textbf{0.953} & \textbf{0.953} & 0.890 & 0.889 & 0.834 & 0.907 & 0.745 & 0.831\\
StarLightCurves & \textbf{0.973} & \textbf{0.973} & 0.971 & 0.964 & 0.968 & 0.967 & 0.949 & 0.907\\
Strawberry & \textbf{0.978} & \textbf{0.978} & 0.965 & 0.954 & 0.951 & 0.965 & 0.916 & 0.941\\
SwedishLeaf & \textbf{0.954} & 0.950 & 0.942 & 0.914 & 0.880 & 0.923 & 0.738 & 0.792\\
Symbols & \textbf{0.979} & \textbf{0.979} & 0.976 & 0.963 & 0.885 & 0.916 & 0.786 & 0.950\\
SyntheticControl & \textbf{1.000} & \textbf{1.000} & 0.997 & 0.987 & \textbf{1.000} & 0.990 & 0.490 & 0.993\\
ToeSegmentation1 & 0.930 & 0.934 & \textbf{0.947} & 0.939 & 0.864 & 0.930 & 0.807 & 0.772\\
ToeSegmentation2 & \textbf{0.923} & 0.915 & 0.915 & 0.900 & 0.831 & 0.877 & 0.615 & 0.838\\
Trace & \textbf{1.000} & \textbf{1.000} & \textbf{1.000} & 0.990 & \textbf{1.000} & \textbf{1.000} & \textbf{1.000} & \textbf{1.000}\\
TwoLeadECG & \textbf{0.999} & 0.998 & 0.987 & \textbf{0.999} & 0.993 & 0.976 & 0.871 & 0.905\\
TwoPatterns & \textbf{1.000} & \textbf{1.000} & \textbf{1.000} & 0.999 & \textbf{1.000} & 0.999 & 0.466 & \textbf{1.000}\\
UWaveGestureLibraryX & \textbf{0.820} & 0.819 & 0.810 & 0.785 & 0.781 & 0.733 & 0.569 & 0.728\\
UWaveGestureLibraryY & \textbf{0.745} & 0.736 & 0.729 & 0.710 & 0.697 & 0.641 & 0.348 & 0.634\\
UWaveGestureLibraryZ & 0.768 & 0.768 & \textbf{0.770} & 0.757 & 0.721 & 0.690 & 0.655 & 0.658\\
UWaveGestureLibraryAll & 0.966 & \textbf{0.967} & 0.934 & 0.896 & 0.903 & 0.692 & 0.475 & 0.892\\
Wafer & \textbf{0.999} & 0.998 & 0.998 & 0.992 & 0.994 & 0.994 & 0.991 & 0.980\\
Wine & \textbf{0.963} & \textbf{0.963} & 0.889 & 0.815 & 0.759 & 0.778 & 0.500 & 0.574\\
WordSynonyms & \textbf{0.715} & 0.704 & 0.704 & 0.691 & 0.630 & 0.531 & 0.422 & 0.649\\
Worms & \textbf{0.766} & 0.753 & 0.701 & 0.727 & 0.623 & 0.753 & 0.455 & 0.584\\
WormsTwoClass & 0.818 & \textbf{0.857} & 0.805 & 0.792 & 0.727 & 0.753 & 0.584 & 0.623\\
Yoga & \textbf{0.937} & 0.869 & 0.887 & 0.837 & 0.812 & 0.791 & 0.830 & 0.837\\
ACSF1 & 0.850 & 0.850 & \textbf{0.910} & 0.900 & 0.730 & 0.730 & 0.760 & 0.640\\
AllGestureWiimoteX & 0.560 & 0.630 & \textbf{0.777} & 0.763 & 0.703 & 0.697 & 0.259 & 0.716\\
AllGestureWiimoteY & 0.623 & 0.686 & \textbf{0.793} & 0.726 & 0.699 & 0.741 & 0.423 & 0.729\\
AllGestureWiimoteZ & 0.633 & 0.629 & \textbf{0.770} & 0.723 & 0.646 & 0.689 & 0.447 & 0.643\\
BME & \textbf{1.000} & \textbf{1.000} & 0.993 & 0.993 & 0.973 & 0.933 & 0.760 & 0.900\\
Chinatown & 0.985 & \textbf{0.988} & 0.968 & 0.951 & 0.977 & 0.983 & 0.936 & 0.957\\
Crop & \textbf{0.766} & \textbf{0.766} & 0.756 & 0.722 & 0.738 & 0.742 & 0.710 & 0.665\\
EOGHorizontalSignal & 0.577 & 0.572 & 0.544 & \textbf{0.605} & 0.442 & 0.401 & 0.373 & 0.503\\
EOGVerticalSignal & 0.459 & 0.459 & \textbf{0.503} & 0.434 & 0.392 & 0.376 & 0.298 & 0.448\\
EthanolLevel & 0.710 & \textbf{0.712} & 0.484 & 0.382 & 0.424 & 0.486 & 0.260 & 0.276\\
FreezerRegularTrain & \textbf{0.998} & 0.996 & 0.986 & 0.956 & 0.991 & 0.989 & 0.922 & 0.899\\
FreezerSmallTrain & \textbf{0.991} & 0.988 & 0.894 & 0.933 & 0.982 & 0.979 & 0.920 & 0.753\\
Fungi & 0.866 & 0.946 & 0.962 & \textbf{1.000} & 0.527 & 0.753 & 0.366 & 0.839\\
GestureMidAirD1 & 0.592 & 0.592 & \textbf{0.631} & 0.608 & 0.431 & 0.369 & 0.208 & 0.569\\
GestureMidAirD2 & 0.459 & 0.492 & 0.515 & 0.546 & 0.362 & 0.254 & 0.138 & \textbf{0.608}\\
GestureMidAirD3 & 0.323 & 0.315 & \textbf{0.346} & 0.285 & 0.292 & 0.177 & 0.154 & 0.323\\
GesturePebbleZ1 & 0.895 & 0.802 & \textbf{0.930} & 0.919 & 0.378 & 0.395 & 0.500 & 0.791\\
GesturePebbleZ2 & \textbf{0.905} & 0.842 & 0.873 & 0.899 & 0.316 & 0.430 & 0.380 & 0.671\\
GunPointAgeSpan & 0.997 & \textbf{1.000} & 0.994 & 0.994 & 0.984 & 0.994 & 0.991 & 0.918\\
GunPointMaleVersusFemale & \textbf{1.000} & \textbf{1.000} & \textbf{1.000} & 0.997 & 0.994 & 0.997 & \textbf{1.000} & 0.997\\
GunPointOldVersusYoung & \textbf{1.000} & \textbf{1.000} & \textbf{1.000} & \textbf{1.000} & \textbf{1.000} & \textbf{1.000} & \textbf{1.000} & 0.838\\
HouseTwenty & \textbf{0.941} & 0.924 & \textbf{0.941} & 0.933 & 0.782 & 0.790 & 0.815 & 0.924\\
InsectEPGRegularTrain & \textbf{1.000} & \textbf{1.000} & \textbf{1.000} & \textbf{1.000} & \textbf{1.000} & \textbf{1.000} & \textbf{1.000} & 0.872\\
InsectEPGSmallTrain & \textbf{1.000} & \textbf{1.000} & \textbf{1.000} & \textbf{1.000} & \textbf{1.000} & \textbf{1.000} & \textbf{1.000} & 0.735\\
MelbournePedestrian & \textbf{0.964} & 0.962 & 0.959 & 0.944 & 0.942 & 0.949 & 0.741 & 0.791\\
MixedShapesRegularTrain & \textbf{0.940} & 0.935 & 0.922 & 0.905 & 0.911 & 0.855 & 0.879 & 0.842\\
MixedShapesSmallTrain & \textbf{0.892} & 0.887 & 0.881 & 0.860 & 0.813 & 0.735 & 0.828 & 0.780\\
PickupGestureWiimoteZ & \textbf{0.820} & \textbf{0.820} & \textbf{0.820} & 0.740 & 0.620 & 0.600 & 0.240 & 0.660\\
PigAirwayPressure & 0.433 & 0.432 & \textbf{0.683} & 0.510 & 0.413 & 0.380 & 0.120 & 0.106\\
PigArtPressure & 0.820 & 0.830 & \textbf{0.966} & 0.928 & 0.808 & 0.524 & 0.774 & 0.245\\
PigCVP & 0.654 & 0.653 & \textbf{0.870} & 0.788 & 0.649 & 0.615 & 0.596 & 0.154\\
PLAID & 0.356 & 0.355 & 0.561 & 0.555 & 0.495 & 0.445 & 0.419 & \textbf{0.840}\\
PowerCons & 0.995 & \textbf{1.000} & 0.972 & 0.900 & 0.933 & 0.961 & 0.911 & 0.878\\
Rock & \textbf{0.760} & \textbf{0.760} & 0.700 & 0.580 & 0.580 & 0.600 & 0.680 & 0.600\\
SemgHandGenderCh2 & 0.939 & 0.944 & \textbf{0.963} & 0.890 & 0.882 & 0.837 & 0.725 & 0.802\\
SemgHandMovementCh2 & 0.833 & 0.836 & \textbf{0.893} & 0.789 & 0.593 & 0.613 & 0.420 & 0.584\\
SemgHandSubjectCh2 & 0.945 & 0.924 & \textbf{0.951} & 0.853 & 0.771 & 0.753 & 0.484 & 0.727\\
ShakeGestureWiimoteZ & 0.920 & 0.920 & \textbf{0.940} & 0.920 & 0.820 & 0.860 & 0.760 & 0.860\\
SmoothSubspace & \textbf{1.000} & \textbf{1.000} & 0.993 & 0.960 & 0.913 & 0.953 & 0.827 & 0.827\\
UMD & \textbf{1.000} & \textbf{1.000} & \textbf{1.000} & 0.993 & 0.993 & 0.986 & 0.910 & 0.993\\
DodgerLoopDay & \textbf{0.675} & \textbf{0.675} & 0.562 &  -- & -- & -- & 0.200 & 0.500\\
DodgerLoopGame & \textbf{0.971} & 0.942 & 0.841 & -- & -- & -- & 0.696 & 0.877\\
DodgerLoopWeekend & \textbf{0.986} & \textbf{0.986} & 0.964 & -- & -- & -- & 0.732 & 0.949\\

    \midrule
    AVG  & \textbf{0.841} & 0.838 & 0.836 & 0.806 & 0.761 & 0.757 & 0.639 & 0.729\\
    Rank &\textbf{ 1.757} & 1.969 & 2.328 & 3.640 & 4.508 & 4.383 & 6.117 & 5.125\\
    \bottomrule 
  \caption{Full results of univariate time series classification on 128 UCR datasets.}\label{tab:app:full-ucr}
\end{longtable}
}

\newpage
\begin{longtable}[t]{lccccccccccc}
  \toprule
    Dataset  & $\text{InfoTS}_s$ & InfoTS & TS2Vec & T-Loss & TNC & TS-TCC & TST & DTW \\
    \midrule
    \endhead
ArticularyWordRecognition& \textbf{0.993}& 0.987 &0.987 & 0.943 & 0.973 & 0.953 & 0.977 & 0.987 \\
AtrialFibrillation& \textbf{0.267} & 0.200  & 0.200 & 0.133 & 0.133 & \textbf{0.267} & 0.067 & 0.200 \\
BasicMotions &\textbf{1.000}& 0.975 & 0.975 & \textbf{1.000} & 0.975 & \textbf{1.000} & 0.975 & 0.975 \\
CharacterTrajectories&0.987& 0.974  & \textbf{0.995} & 0.993 & 0.967 & 0.985 & 0.975 & 0.989 \\
Cricket & \textbf{1.000} &  0.986 & 0.972 & 0.972 & 0.958 & 0.917 & \textbf{1.000} & \textbf{1.000} \\
DuckDuckGeese & 0.600 & 0.540 & \textbf{0.680} & 0.650 & 0.460 & 0.380 & 0.620 & 0.600 \\
                     EigenWorms & 0.748 & 0.733 & \textbf{0.847 }& 0.840 & 0.840 & 0.779 & 0.748 & 0.618 \\
Epilepsy & \textbf{0.993} & 0.971 & 0.964 & 0.971 & 0.957 & 0.957 & 0.949 & 0.964 \\
ERing & \textbf{0.953}& 0.949 & 0.874 & 0.133 & 0.852 & 0.904 & 0.874 & 0.133 \\
EthanolConcentration& \textbf{0.323} & 0.281 & 0.308 & 0.205 & 0.297 & 0.285 & 0.262 & \textbf{0.323} \\
FaceDetection  & 0.525& 0.534 & 0.501 & 0.513 & 0.536 & \textbf{0.544} & 0.534 & 0.529 \\
FingerMovements & 0.620 &\textbf{0.630} & 0.480 & 0.580 & 0.470 & 0.460 & 0.560 & 0.530 \\
HandMovementDirection&  \textbf{0.514} &0.392 & 0.338 & 0.351 & 0.324 & 0.243 & 0.243 & 0.231 \\
Handwriting &\textbf{0.554}& 0.452 & 0.515 & 0.451 & 0.249 & 0.498 & 0.225 & 0.286 \\
Heartbeat & \textbf{0.771}& 0.722 & 0.683 & 0.741 & 0.746 & 0.751 & 0.746 & 0.717 \\
JapaneseVowels & 0.986& 0.984 & 0.984 & \textbf{0.989} & 0.978 & 0.930 & 0.978 & 0.949 \\
Libras &\textbf{0.889} & 0.883 & 0.867 & 0.883 & 0.817 & 0.822 & 0.656 & 0.870 \\
LSST & 0.593 & 0.591 & 0.537 & 0.509 & \textbf{0.595} & 0.474 & 0.408 & 0.551 \\
MotorImagery &0.610 & \textbf{0.630} & 0.510 & 0.580 & 0.500 & 0.610 & 0.500 & 0.500 \\
NATOPS & \textbf{0.939} & 0.933 & 0.928 & 0.917 & 0.911 & 0.822 & 0.850 & 0.883 \\
PEMS-SF & \textbf{0.757}& 0.751 & 0.682 & 0.676 & 0.699 & 0.734 & 0.740 & 0.711 \\
PenDigits & 0.989 & \textbf{0.990} & 0.989 & 0.981 & 0.979 & 0.974 & 0.560 & 0.977 \\
PhonemeSpectra & 0.233& 0.249 & 0.233 & 0.222 & 0.207 & \textbf{0.252} & 0.085 & 0.151 \\
RacketSports & 0.829&  \textbf{0.855} & \textbf{0.855} & \textbf{0.855} & 0.776 & 0.816 & 0.809 & 0.803 \\
SelfRegulationSCP1 & \textbf{0.887}& 0.874 & 0.812 & 0.843 & 0.799 & 0.823 & 0.754 & 0.775 \\
SelfRegulationSCP2 &0.572 & \textbf{0.578} & \textbf{0.578} & 0.539 & 0.550 & 0.533 & 0.550 & 0.539 \\
SpokenArabicDigits & 0.932 & 0.947 & \textbf{0.988} & 0.905 & 0.934 & 0.970 & 0.923 & 0.963 \\
StandWalkJump & \textbf{0.467} & \textbf{ 0.467 }& \textbf{0.467} & 0.333 & 0.400 & 0.333 & 0.267 & 0.200 \\
UWaveGestureLibrary & 0.884 & 0.884 & \textbf{0.906} & 0.875 & 0.759 & 0.753 & 0.575 & 0.903 \\
InsectWingbeat & \textbf{0.472}& 0.470 & 0.466 & 0.156 & 0.469 & 0.264 & 0.105 & -- \\
    \midrule
    Avg. ACC & \textbf{0.730}& 0.714& 0.704& 0.658 & 0.670 & 0.668 & 0.617 & 0.629 \\
    Avg. Rank & \textbf{1.967}& 2.633& 3.067& 3.833 & 4.367 & 4.167 & 5.0 & 4.366 \\
    \bottomrule
  \caption{Full results of multivariate time series classification on 30 UEA datasets.
  }\label{tab:app:full-uea}
\end{longtable}

\end{document}